\documentclass[sigconf]{acmart}

\copyrightyear{2022}
\acmYear{2022}
\setcopyright{rightsretained}

\acmConference[KDD '22]{Proceedings of the 28th ACM SIGKDD Conference on Knowledge Discovery and Data Mining}{August 14--18, 2022}{Washington, DC, USA}
\acmBooktitle{Proceedings of the 28th ACM SIGKDD Conference on Knowledge Discovery and Data Mining (KDD '22), August 14--18, 2022, Washington, DC, USA}
\acmISBN{978-1-4503-9385-0/22/08}
\acmDOI{10.1145/3534678.3539265}

\usepackage{etoolbox}
\makeatletter
\patchcmd{\maketitle}{\@copyrightpermission}{
   \begin{minipage}{0.3\columnwidth}
     \href{https://creativecommons.org/licenses/by/4.0/}{\includegraphics[width=0.90\textwidth]{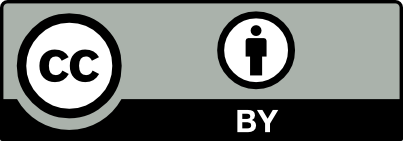}}
   \end{minipage}\hfill
   \begin{minipage}{0.7\columnwidth}
     \href{https://creativecommons.org/licenses/by/4.0/}{This work is licensed under a Creative Commons Attribution International 4.0 License.}
   \end{minipage}

   \vspace{5pt}
}{}{}

\makeatother

\AtBeginDocument{%
	\providecommand\BibTeX{{%
			\normalfont B\kern-0.5em{\scshape i\kern-0.25em b}\kern-0.8em\TeX}}}
			
\usepackage{flushend}
\usepackage{balance}
\usepackage{lineno}
\usepackage{amsmath,amsfonts}
\usepackage{algorithm}
\usepackage{algorithmic}
\usepackage{graphicx}
\usepackage{textcomp}
\usepackage{xcolor}

\usepackage{amsthm}
\usepackage{url}
\usepackage{float}
\usepackage{multirow}
\usepackage{multicol}
\usepackage{color}
\usepackage{bm}
\usepackage{bbm}
\usepackage{caption}
\usepackage{subcaption}
\usepackage{appendix}

\newtheorem{definition}{Definition}

\usepackage{pifont}

\newcommand{\X}{\mathbf{X}}

\usepackage{array}
\newcolumntype{L}[1]{>{\raggedright\let\newline\\\arraybackslash\hspace{0pt}}m{#1}}
\newcolumntype{C}[1]{>{\centering\let\newline  \\\arraybackslash\hspace{0pt}}m{#1}}
\newcolumntype{R}[1]{>{\raggedleft\let\newline \\\arraybackslash\hspace{0pt}}m{#1}}

\renewcommand{\shortauthors}{Song Wang et al.}

\usepackage{ bbold }
\DeclareMathOperator*{\argmax}{argmax}

\begin{document}

	\title{Task-Adaptive Few-shot Node Classification}
	
	\author{Song Wang}
\affiliation{%
  \institution{University of Virginia} \country{}}
\email{sw3wv@virginia.edu}

		\author{Kaize Ding}
\affiliation{%
  \institution{Arizona State University}  \country{}}

\email{kding9@asu.edu}

		\author{Chuxu Zhang}
\affiliation{%
  \institution{Brandeis University} \country{}}
\email{chuxuzhang@brandeis.edu}

				\author{Chen Chen}
\affiliation{%
  \institution{University of Virginia} \country{}}
\email{zrh6du@virginia.edu}

					\author{Jundong Li}
\affiliation{%
  \institution{University of Virginia} \country{}}
\email{jundong@virginia.edu}

	\begin{abstract}
		Node classification is of great importance among various graph mining tasks. In practice, real-world graphs generally follow the long-tail distribution, where a large number of classes only consist of limited labeled nodes. Although Graph Neural Networks (GNNs) have achieved significant improvements in node classification, their performance decreases substantially in such a few-shot scenario. The main reason can be attributed to the vast generalization gap between meta-training and meta-test due to the task variance caused by different node/class distributions in meta-tasks (i.e., node-level and class-level variance). 
		Therefore, to effectively alleviate the impact of task variance, we propose a task-adaptive node classification framework under the few-shot learning setting. Specifically, we first accumulate meta-knowledge across classes with abundant labeled nodes. Then we transfer such knowledge to the classes with limited labeled nodes via our proposed task-adaptive modules. In particular, to accommodate the different node/class distributions among meta-tasks, we propose three essential modules to perform \emph{node-level}, \emph{class-level}, and \emph{task-level} adaptations in each meta-task, respectively. In this way, our framework can conduct adaptations to different meta-tasks and thus advance the model generalization performance on meta-test tasks. Extensive experiments on four prevalent node classification datasets demonstrate the superiority of our framework over the state-of-the-art baselines. Our code is provided at \href{https://github.com/SongW-SW/TENT}{https://github.com/SongW-SW/TENT}.
		
	\end{abstract}

	\maketitle

	\section{INTRODUCTION}
	Recently, extensive research efforts have been devoted to the node classification task, which aims at predicting class labels for unlabeled nodes in a graph. In real-world scenarios, the task of node classification yields an expansive variety of practical applications~\cite{mcauley2012learning,tang2008arnetminer}. For example, predicting chemical properties for proteins in a protein network is an important problem in bioinformatics~\cite{szklarczyk2019string}, which can be formulated as the node classification problem.
	In recent years, the state-of-the-art approaches for node classification often utilize Graph Neural Networks (GNNs)~\cite{wu2020comprehensive,velivckovic2017graph,xu2018powerful} in a semi-supervised manner~\cite{kipf2017semi}. Specifically, for each node, GNNs aim to learn a vector representation for each node by transforming and aggregating information from its neighbors. The learned representations will be further utilized for the classification task in an end-to-end manner. 
	Nevertheless, these approaches typically require sufficient labeled nodes for all classes in achieving a decent classification performance~\cite{zhou2019meta}. 
	In practice, although we can access a large number of labeled nodes for certain classes, many other classes may only contain a limited number of labeled nodes. Here we refer to the former classes as \emph{base classes} and the latter as \emph{novel classes}.
	For example, in a protein network~\cite{hu2020open}, newly discovered chemical properties with limited protein nodes are considered as novel classes, while common properties with abundant protein nodes are considered as base classes.
	Due to the widespread existence of novel classes in real-world graphs,
	many recent studies~\cite{ding2020graph,wang21AMM,liu2021relative} focus on the problem of classifying nodes in novel classes, known as the \emph{few-shot node classification} problem.

	
				\begin{figure}[t]
					
	    \centering
	    \includegraphics[width=0.99\linewidth]{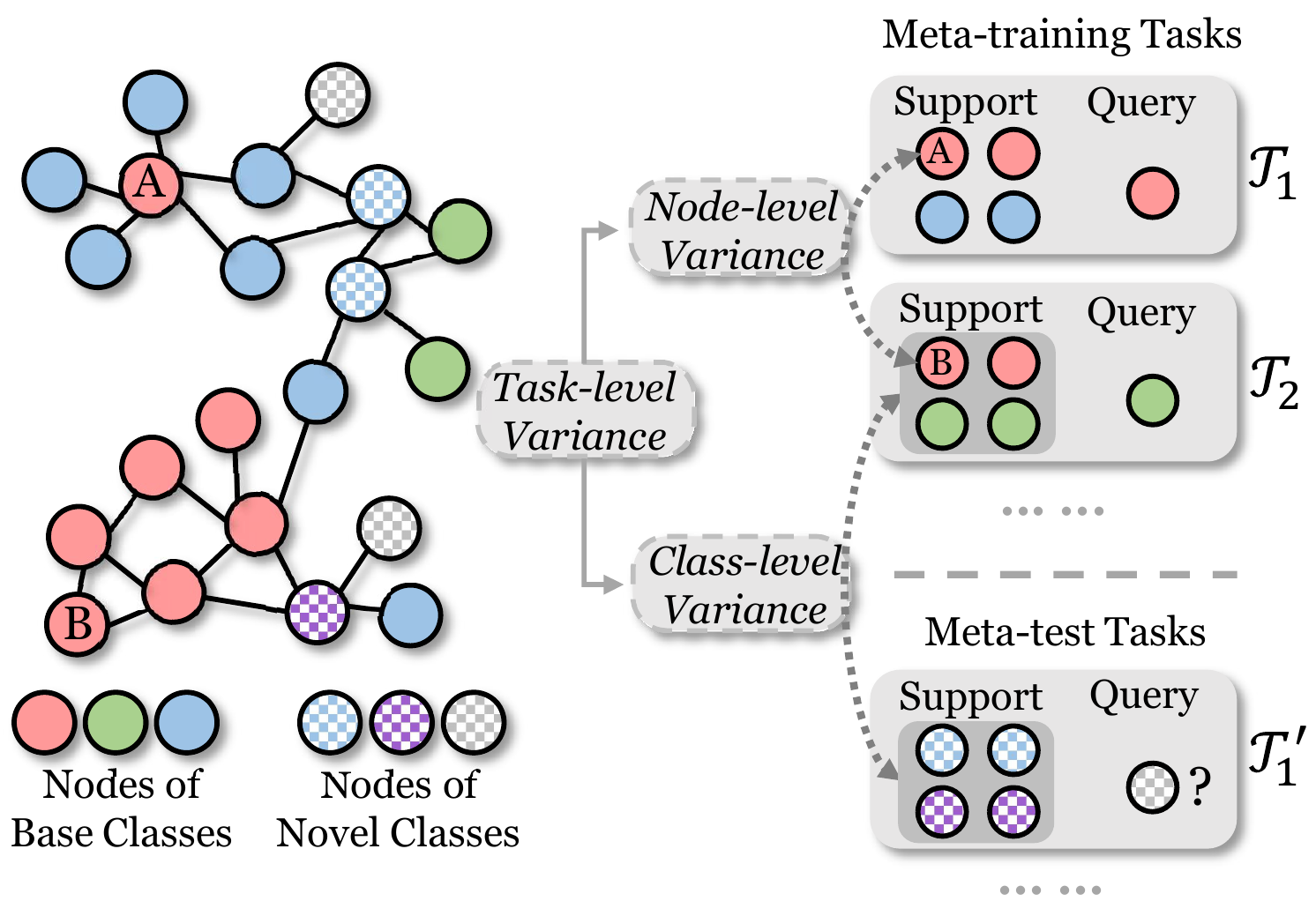}
	    \vspace{-2.5mm}
	    \caption{Issues of task variance of existing few-shot node classification frameworks.}
	    \vspace{-5.5mm}
	    \label{fig:intro}
	  
	\end{figure}

	To tackle the few-shot node classification problem, recent works typically
	strive to extract transferable knowledge from base classes
	and then generalize such knowledge to novel classes~\cite{zhou2019meta,yao2020graph,ding2020inductive}.
	More specifically, these works learn from base classes across a series of \emph{meta-training} tasks and evaluate the model on \emph{meta-test} tasks sampled from novel classes (we refer to both meta-training and meta-test tasks as meta-tasks). In fact, each meta-task 
	contains a small number of \emph{support nodes} as references and several \emph{query nodes} to be classified. 
	Since support nodes and query nodes in each meta-task are sampled from nodes on the entire graph,
	there could exist large variance among different meta-tasks (i.e., task variance)~\cite{huang2020graph}. Therefore, the crucial part of few-shot node classification is to ensure that the underlying model has the generalization capability to handle a variety of meta-tasks in the presence of massive task variance~\cite{lichtenstein2020tafssl,suo2020tadanet}. However, despite much progress has been made in few-shot node classification, recent studies ignore the task variance and treat each meta-task identically~\cite{ding2020graph,wang21AMM,liu2021relative}. As a result, the task variance significantly jeopardizes the model generalization capability to meta-test tasks even when the performance on meta-training tasks is satisfactory~\cite{oreshkin2018tadam}.

	Despite the importance of considering task variance, reducing its adverse impact remains non-trivial. In essence, there are two main factors that constitute such task variance. 
	First, the \emph{Node-level Variance} widely exists among meta-tasks and can lead to task variance. Specifically, node-level variance 
	represents the differences of node features and local structures of nodes across different meta-tasks.
	For example, in addition to the common difference in node features, the red class nodes $A$ and $B$ in Fig.~\ref{fig:intro} also have different connectivity patterns in terms of neighboring nodes (node $A$ is surrounded by blue nodes, while node $B$ is only connected to red nodes). It should be noted that few-shot node classification models generally learn crucial information from the support nodes within each meta-task to perform classification on the query nodes. Therefore, if the variance among the support nodes is too large, it will become difficult to extract decisive information for classification. In other words, it is vital to consider node-level variance for the purpose of handling task variance.
    Second, \emph{Class-level Variance} may also cause task variance. Class-level variance denotes the difference in class distributions among meta-tasks. 
    In practice, since many real-world graphs contain a large number of node classes, the distribution of classes in each meta-task varies greatly~\cite{zhou2019meta,wang21AMM}. For example, in Fig.~\ref{fig:intro}, different meta-tasks consist of a variety of classes (e.g., red and blue classes in $\mathcal{T}_1$ and dotted blue and green classes in $\mathcal{T}'_1$). Since the model evaluation is conducted on a vast number of meta-test tasks, the model will encounter many distinct classes during meta-test. That being said, in the presence of massive class-level variance, the resulting task variance will substantially deteriorate the generalization performance on meta-test tasks.

	To alleviate the adverse impact of task variance resulting from the above two factors (i.e., node-level and class-level variance), we propose a novel \textbf{\underline{T}}ask-adaptiv\textbf{\underline{E}} few-shot \textbf{\underline{N}}ode classifica\textbf{\underline{T}}ion framework, named as \textsc{TENT}. Specifically, we aim to alleviate task variance via performing task adaptations from three perspectives.
	First, to handle node-level variance, we perform node-level adaptations via constructing a class-ego subgraph for each class in each meta-task. Specifically, such a subgraph explicitly connects nodes in the same class and their neighbors with a virtual class node. 
	In this way, the neighbors of nodes in the same class are aggregated in this subgraph to reduce the influence of node-level variance.
	Second, to deal with class-level variance, we design a class-specific GNN to leverage information from different classes and perform class-level adaptations. Third, 
	to reduce the adverse impact of task variance during classification on query nodes, we propose to perform task-level adaptations via maximally preserving the mutual information between query nodes and support nodes in each meta-task. As a result, our proposed framework can conduct classification in a task-adaptive manner to alleviate the adverse impact of task variance. In summary, our main contributions are three-folds:
	\vspace{-0.0in}
	\begin{itemize}
	\setlength{\itemsep}{0.06in}
	    \item \textbf{Problem.} We investigate the limitations of existing few-shot node classification methods from the lens of task variance and discuss the importance and necessity of task adaptations for few-shot node classification.
	    \item \textbf{Method.} We develop a novel task-adaptive few-shot node classification framework with three essential modules: (1) \emph{node-level adaptation} to mitigate node-level variance; (2) \emph{class-level adaptation} to alleviate the problem of class-level variance; and
	    (3) \emph{task-level} adaptation to consider task variance during classification on query nodes.
	    \item \textbf{Experiments.} We conduct experiments on four benchmark node classification datasets under the few-shot setting and demonstrate the superiority of our proposed framework.
	\end{itemize}

		\begin{figure*}[htbp]
	    \centering
	    \includegraphics[width=0.95\textwidth]{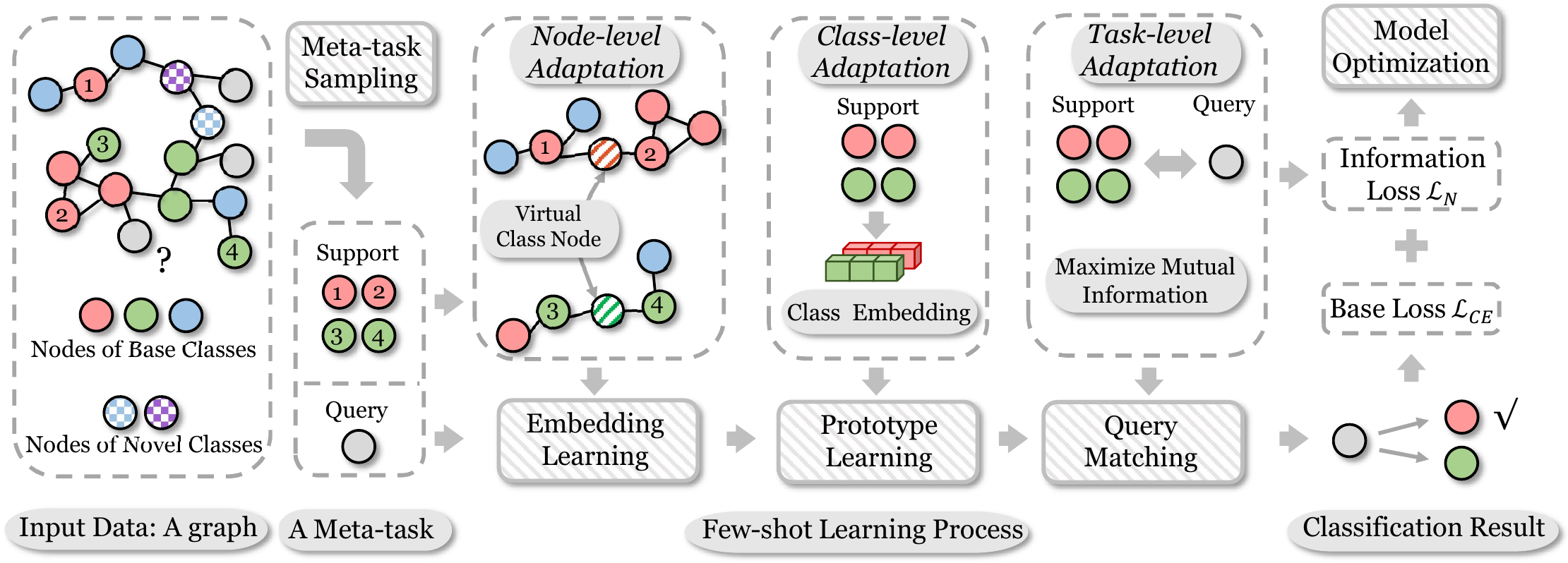}
	    \vspace{-0.1in}
	    \caption{An illustration of the overall process of TENT. We first sample a meta-task from the given graph. Then we construct subgraphs for node-level adaptions and utilize node embeddings in each class for class-level adaptations. We further maximize the mutual information between the support set and the query set during query matching for task-level adaptations. }

	    \label{fig:model}
	    	    \vspace{-0.1in}
	\end{figure*}

	\section{Preliminaries}
	
	\subsection{Problem Statement}
	Formally, let $G=(\mathcal{V},\mathcal{E},\X)$ denote an attributed graph, where $\mathcal{V}$ is the set of nodes, $\mathcal{E}$ is the set of edges, and $\X\in\mathbb{R}^{|\mathcal{V}|\times d}$ is the feature matrix of nodes with $d$ denoting the feature dimension. Moreover, we denote the entire set of node classes as $\mathcal{C}$, which can be further divided into two categories: $\mathcal{C}_b$ and $\mathcal{C}_n$, where $\mathcal{C}=\mathcal{C}_b\cup\mathcal{C}_n$ and $\mathcal{C}_b\cap\mathcal{C}_n=\emptyset$. Here $\mathcal{C}_b$ and $\mathcal{C}_n$ denote the sets of base and novel classes, respectively. It is worth mentioning that the number of labeled nodes in $\mathcal{C}_b$ is sufficient, while it is typically small in $\mathcal{C}_n$~\cite{zhou2019meta,ding2020graph,liu2021relative}. Then we can formulate the studied problem of few-shot node classification as follows:
	
	\begin{definition}
	\textbf{Few-shot Node Classification:} Given an attributed graph $G=(\mathcal{V},\mathcal{E},\X)$, our goal is to develop a machine learning model such that after training on labeled nodes in $\mathcal{C}_b$, the model can accurately predict labels for the nodes (i.e., query set $\mathcal{Q}$) in $\mathcal{C}_n$ with only a limited number of labeled nodes (i.e., support set $\mathcal{S}$).
	\end{definition}
	
	More specifically, if the support set $\mathcal{S}$ contains exactly $K$ nodes for each of $N$ classes from $\mathcal{C}_n$, and the query set $\mathcal{Q}$ are sampled from these $N$ classes, the problem is called $N$-way $K$-shot node classification. Essentially, the objective of few-shot node classification is to learn a classifier that can be fast adapted to $\mathcal{C}_n$ with only limited labeled nodes. Thus, the crucial part is to learn transferable knowledge from $\mathcal{C}_b$ and generalize it to $\mathcal{C}_n$.
	
	\subsection{Episodic Learning}
In practice, we adopt the episodic learning framework for both meta-training and meta-test, which has proven to be effective in many areas~\cite{snell2017prototypical,finn2017model,vinyals2016matching,xiong2018one,ding2020graph}. Specifically, the meta-training and meta-test processes are conducted on a certain number of \emph{meta-training tasks} and \emph{meta-test tasks}, respectively. These meta-tasks share a similar structure, except that meta-training tasks are sampled from $\mathcal{C}_b$, while meta-test tasks are sampled from $\mathcal{C}_n$. The main idea of few-shot node classification is to keep the consistency between meta-training and meta-test to improve the generalization performance. 
	 
	To construct a meta-training (or meta-test) task $\mathcal{T}_t$, we first randomly sample $N$ classes from $\mathcal{C}_b$ (or $\mathcal{C}_n$). Then we randomly sample $K$ nodes from each of the $N$ classes (i.e., $N$-way $K$-shot) to establish the support set $\mathcal{S}_t$. Similarly, the query set $\mathcal{Q}_t$ consists of $Q$ different nodes (distinct from $\mathcal{S}_t$) from the same $N$ classes. The components of the sampled meta-task $\mathcal{T}_t$ can be denoted as follows: 
	\begin{equation}
	    \begin{aligned}
	    \mathcal{S}_t&=\{(v_1,y_1),(v_2,y_2),\dotsc,(v_{N\times K},y_{N\times K})\},\\
	    \mathcal{Q}_t&=\{(q_1,y'_1),(q_2,y'_2),\dotsc,(q_{Q},y'_{Q})\},\\
	    \mathcal{T}_t&=\{\mathcal{S}_t,\mathcal{Q}_t\},
	    \end{aligned}
	\end{equation}
	where $v_i$ (or $q_i)$ is a node in $\mathcal{V}$, and $y_i$ (or $y'_i$) is the corresponding label. In this way, the whole training process is conducted on a set of $T$ meta-training tasks $\mathcal{T}_{train}=\{\mathcal{T}_t\}_{t=1}^T$. After training, the model has learned the transferable knowledge from $\mathcal{T}_{train}$ and will generalize it to meta-test tasks $\mathcal{T}_{test}=\{\mathcal{T}'_t\}_{t=1}^{T_{test}}$ sampled from $\mathcal{C}_n$.

\section{Our Proposed Framework}
	
In this section, we introduce the overall structure of our proposed
framework \textsc{TENT} in detail. As illustrated in Fig.~\ref{fig:model}, we formulate the \emph{few-shot node classification} problem under the prevailing $N$-way $K$-shot learning framework, which means a meta-task contains $K$ nodes for each of $N$ classes as the support set. In addition, the query set consists of $Q$ unlabeled nodes to be classified from these $N$ classes. Specifically, our framework follows the prevalent three phases for few-shot learning: embedding learning, prototype learning, and query matching. Generally, in each meta-task, we learn embeddings for its nodes and then learn a prototype (i.e., embedding of a class in the support set) based on the node embeddings. Finally, the model matches query nodes with these prototypes via specific matching functions to output classification results. Nevertheless, these three steps ignore the task variance that widely exists among meta-tasks. Therefore, as illustrated in Fig.~\ref{fig:model}, we propose to perform three levels of adaptations (node-level, class-level, and task-level adaptations) in these three phases, respectively, to alleviate the adverse impact of task variance.

	    \vspace{-0.1in}
	\subsection{Node-level Adaptation}
During the embedding learning phase, existing methods learn embeddings for nodes in each meta-task from the entire graph~\cite{zhou2019meta,ding2020graph,wang21AMM}. Since nodes are distributed across the entire graph, the learned node representations can be easily influenced by node-level variance (i.e., have different connectivity patterns in terms of neighboring nodes). Instead, we perform node-level adaptations in each meta-task, which aims to modify the neighbors of support nodes to reduce node-level variance caused by different connectivity patterns. Toward this goal, we explicitly construct a subgraph for each class in each meta-task via a virtual class node, which connects to the $K$ support nodes in that class. In addition, we also include the one-hop neighbors of these $K$ nodes in this subgraph. 
By doing the above, we can aggregate local structures of support nodes in the same class into this subgraph, which contains the virtual class node, $K$ support nodes, and one-hop neighbors of these $K$ nodes. Here the virtual class node acts as a bridge to explicitly connect these $K$ support nodes and their one-hop neighbors that can be originally far from each other on the graph. Moreover, its embedding will be used as the prototype of this class since it is the centroid node of the subgraph. As a result, in this subgraph, 
support nodes will share a similar neighbor node set because the neighboring nodes of support nodes are explicitly connected.
Since the neighbors of support nodes become more similar in this subgraph, we can effectively reduce the node-level variance. 
	 We denote the constructed subgraphs in each meta-task as \emph{class-ego subgraphs}.
	
	Specifically, given a meta-task $\mathcal{T}$ and its support set $\mathcal{S}$ ($|\mathcal{S}|=N\times K$) on a graph $G=(\mathcal{V},\mathcal{E},\X)$, we aim to construct a class-ego subgraph for each of the $N$ classes in $\mathcal{T}$. Before the construction of class-ego subgraphs, we employ a GNN~\cite{wu2020comprehensive,kipf2017semi} parameterized by $\phi$ to perform message propagation on the entire graph $G$ and generate first-step node representations for nodes in $\mathcal{V}$ as follows:
	\begin{equation}
	    \mathbf{H}=\text{GNN}_\phi(\mathcal{V},\mathcal{E},\X),
	    \label{eq:first-step emb}
	\end{equation}
	where $\mathbf{H}\in\mathbb{R}^{|\mathcal{V}|\times d_h}$ denotes the first-step representations of nodes in $\mathcal{V}$ and $d_h$ is the output dimension of $\text{GNN}_\phi$. In this way, $\mathbf{H}$ will act as the input node representations for the class-ego subgraphs. 
	
	Let $\mathcal{S}_i$ denote the set of nodes belonging to the $i$-th class in $\mathcal{S}$, which means $|\mathcal{S}_i|=K$, $i=1,2,\dotsc,N$. To construct the class-ego subgraph from these nodes, we first create a virtual class node $c_i$ and connect it to all nodes in $\mathcal{S}_i$. Then to incorporate the local graph structures, we also extract all one-hop neighbors of nodes in $\mathcal{S}_i$ to establish a neighbor node set $\mathcal{N}_i=\bigcup_{j=1}^K\mathcal{N}_i^j$, where $\mathcal{N}_i^j$ denotes the set of neighbors of the $j$-th node in $\mathcal{S}_i$. In this way, the final node set of the class-ego subgraph is aggregated as $\mathcal{V}_i=\{c_i\}\cup\mathcal{S}_i\cup\mathcal{N}_i$. After that, we accordingly denote the extracted edge set of nodes in $\mathcal{V}_i$ as $\mathcal{E}_i$. To obtain the input node features for $\mathcal{V}_i$, we utilize the corresponding first-step representations from $\mathbf{H}$. However, we still need to compute the representation of $c_i$ since it is newly created. Here we propose to initiate its representation $\mathbf{h}_{c_i}$ as follows:
	\begin{equation}
	    \mathbf{h}_{c_i}=\text{MEAN}(\mathbf{h}_v|v\in\mathcal{S}_i),
	\end{equation}
	where $\mathbf{h}_{c_i}\in\mathbb{R}^{d_h}$ and $\mathbf{h}_v$ is the first-step node representation of node $v$. $\text{MEAN}$ denotes the averaging operation.
	In this way, the input node features for $\mathcal{V}_i$ can be obtained as $\mathbf{X}_i$. Then the class-ego subgraph can be constructed and denoted as $G_i=(\mathcal{V}_i,\mathcal{E}_i,\X_i)$. 
	As a result, we can achieve node-level adaptations by learning node representations on the subgraphs with reduced node-level variance.
	
	\subsection{Class-level Adaptation}
	Typically, after learning the node representations in $\mathcal{T}$, existing models learn prototypes for classes in $\mathcal{T}$ by aggregating node representations in the same class~\cite{ding2020graph,snell2017prototypical}. However, this strategy can be easily influenced by class-level variance and thus renders suboptimal generalization performance since it treats each class identically.
	Instead, we propose to perform class-level adaptations, which aim to obtain prototypes for classes in a class-adaptive manner. In particular, we design a class-specific adapter to adjust GNN parameters regarding different classes in $\mathcal{T}$. In this way, our framework can leverage the 
	discriminative information in each class for class-level adaptations and reduce the adverse impact of class-level variance.
	
	Specifically, we use a new $\text{GNN}_\theta$ parameterized by $\theta$ on the class-ego subgraphs to learn prototypes.
	Then we adapt $\theta$ according to the first-step representations of nodes in each class (i.e., $\left\{\mathbf{h}_v|v\in\mathcal{S}_i\right\}, i=1,2,\dotsc,N$) in meta-task $\mathcal{T}$. To comprehensively incorporate the information in each class, we leverage the feature-wise linear modulations~\cite{perez2018film,wen2021meta} to perform class-level adaptations:
	\begin{equation}
\alpha_i=\text{MLP}_\alpha \left(\text{MEAN}\left(\left\{\mathbf{h}_v|v\in\mathcal{S}_i\right\}\right)\right),
	\end{equation}
	\begin{equation}
\beta_i=\text{MLP}_\beta \left(\text{MEAN}\left(\left\{\mathbf{h}_v|v\in\mathcal{S}_i\right\}\right)\right),
\end{equation}
	where $\mathcal{S}_i$ is the set of nodes belonging to the $i$-th class in $\mathcal{S}$. $\alpha_i\in\mathbb{R}^{d_\theta}$ and $\beta_i\in\mathbb{R}^{d_\theta}$ are $d_\theta$-dimensional adaptation parameters, where $d_\theta$ is the total number of parameters in $\text{GNN}_\theta$. With the adaptation parameters $\alpha_i$ and $\beta_i$, we can perform an adaptation based on each class to obtain class-specific GNN parameters as follows:
	\begin{equation}
	    \theta_i=(\alpha_i+\mathbf{1})\circ\theta+\beta_i,
	    \label{eq:adapt_class}
	\end{equation}
	where $\circ$ denotes the element-wise multiplication and $\mathbf{1}$ is a vector of ones to limit the scaling range around one. $\theta_i$ denotes the adapted GNN parameters for the $i$-th class in $\mathcal{S}$. Then we perform message propagation on each class-ego subgraph with the adapted $\theta_i$:
	\begin{equation}
	    \mathbf{s}_i=\text{Centroid}\left(\text{GNN}_{\theta_i}(\mathcal{V}_i,\mathcal{E}_i,\mathbf{X}_i)\right),
	    \label{eq:subgraph_representation}
	\end{equation}
	where $\mathbf{s}_i\in\mathbb{R}^{d_s}$ denotes the learned embedding of the virtual class node (i.e., the centroid node) and acts as the prototype of the $i$-th class. $\text{Centroid}(\cdot)$ denotes the operation of extracting  the centroid node representation from the GNN output. $d_s$ is the output dimension of $\text{GNN}_\theta$.
	As a result, the GNN parameters can absorb the information from each class to reduce the adverse impact of class-level variance.
	Similarly, for the representations of query nodes, we also apply the proposed class-specific adapter. Since labels of query nodes are unknown during meta-training, we utilize the entire support set $\mathcal{S}$ to conduct adaptations for query nodes:
		\begin{equation}
\alpha_q=\text{MLP}_\alpha \left(\text{MEAN}\left(\left\{\mathbf{h}_v|v\in\mathcal{S}\right\}\right)\right),
	\end{equation}
	\begin{equation}
\beta_q=\text{MLP}_\beta \left(\text{MEAN}\left(\left\{\mathbf{h}_v|v\in\mathcal{S}\right\}\right)\right),
\end{equation}
	\begin{equation}
	    \theta_q=(\alpha_q+\mathbf{1})\circ\theta+\beta_q,
	    \label{eq:adapt}
	\end{equation}
where $\theta_q$ is the adapted GNN parameters for query nodes. To obtain the representation $\mathbf{q}_i$ for the $i$-th query node $q_i$ in the query set $\mathcal{Q}$, we extract the 2-hop neighbors of $q_i$ and obtain the corresponding node set $\mathcal{V}^q_i$, edge set $\mathcal{E}^q_i$, and input node features $\mathbf{X}^q_i$. The reason is that using subgraph structures (e.g., considering 2-hop neighbors) to learn node representations provides a more robust generalization ability~\cite{huang2020graph}. Then we utilize the adapted $\text{GNN}_{\theta_q}$ to compute $\mathbf{q}_i$:
	\begin{equation}
	    	    \mathbf{q}_i=\text{Centroid}\left(\text{GNN}_{\theta_q}(\mathcal{V}^q_i,\mathcal{E}^q_i,\mathbf{X}^q_i)\right),
	\end{equation}
	where $\mathbf{q}_i\in\mathbb{R}^{d_s}$ is the embedding of $q_i$ (i.e., the centroid node).
	
	
	\subsection{Task-level Adaptation}
	Although we have achieved node-level and class-level adaptations in the embedding learning and prototype learning phases, respectively, the task variance caused by differences in the support set among meta-tasks still exist in the final query matching phase~\cite{huang2020graph}. Nevertheless, existing methods typically leverage the Euclidean distance metric~\cite{ding2020graph} or an MLP layer~\cite{liu2021relative} to classify query nodes, which ignores the task variance. Instead,
    we propose to perform task-level adaptations in each meta-task, which aim to further reduce the adverse impact of task variance in this phase.
    Specifically, we propose a task-adaptive matching strategy
    to maximally preserve the mutual information between learned representations of nodes in the query set $\mathcal{Q}$ and the support set $\mathcal{S}$.
    As a result, the matching phase on query nodes can incorporate information from the entire support set $\mathcal{S}$ for task-level adaptations.
    
    The optimization problem of maximizing the mutual information can be formulated as follows:
	\begin{equation}
	    \max_{\widetilde{\theta}} I(\mathbf{Q};\mathbf{S})= \max_{\widetilde{\theta}}\sum\limits_{i=1}^Q\sum\limits_{j=1}^Np(q_i,s_j;\widetilde{\theta})\log\frac{p(q_i|s_j;\widetilde{\theta})}{p(q_i;\widetilde{\theta})},
	\end{equation}
	where $\mathbf{Q}\in\mathbb{R}^{Q\times d_s}$ and $\mathbf{S}\in\mathbb{R}^{N\times d_s}$ are learned representations of query nodes in $\mathcal{Q}$ and classes (i.e., prototypes) in $\mathcal{S}$, respectively.
	$Q=|\mathcal{Q}|$ and $N$ is the number of classes in $\mathcal{S}$. $\widetilde{\theta}$ denotes the parameters of our framework to be optimized. $q_i$ is the $i$-th query node in $\mathcal{Q}$ and $s_j$ is the $j$-th class in $\mathcal{S}$. Since the mutual information $I(\mathbf{Q};\mathbf{S})$ is difficult to obtain and thus infeasible to maximize~\cite{oord2018representation}, we re-write the function to obtain an accessible form:
	\begin{equation}
	    I(\mathbf{Q};\mathbf{S})=\sum\limits_{i=1}^Q\sum\limits_{j=1}^Np(q_i|s_j;\widetilde{\theta})p(s_j;\widetilde{\theta})\log\frac{p(q_i|s_j;\widetilde{\theta})}{p(q_i;\widetilde{\theta})}.
	\end{equation}
	Since each meta-task contains $N$ classes, we may assume that the prior probability of $p(s_j;\widetilde{\theta})$ follows a uniform distribution
	and set it as $p(s_j;\widetilde{\theta})=1/N$.
	Since $p(s_j;\widetilde{\theta})$ is a constant, according to the Bayes' theorem, the objective function becomes:
	\begin{equation}
	    \begin{aligned}
	        I(\mathbf{Q};\mathbf{S})
	        &=\frac{1}{N}\sum\limits_{i=1}^Q\sum\limits_{j=1}^Np(q_i|s_j;\widetilde{\theta})\log\frac{p(s_j|q_i;\widetilde{\theta})}{p(s_j;\widetilde{\theta})}\\
	        &=\frac{1}{N}\sum\limits_{i=1}^Q\sum\limits_{j=1}^Np(q_i|s_j;\widetilde{\theta})\left(\log(p(s_j|q_i;\widetilde{\theta}))-\log\left(\frac{1}{N}\right)\right).
	    \end{aligned}
	\end{equation}
	To further estimate $p(q_i|s_j;\widetilde{\theta})$, we compute it by $p(q_i|s_j;\widetilde{\theta})=\mathbb{1}(q_i\in s_j)$, where $\mathbb{1}(q_i\in s_j)=1$ if $q_i$ belongs to the class represented by $s_j$; otherwise $\mathbb{1}(q_i\in s_j)=0$. In this way, the above objective function is simplified as follows:
	\begin{equation}
	    I(\mathbf{Q};\mathbf{S})=\frac{1}{N}\sum\limits_{i=1}^Q\sum\limits_{j=1}^N\mathbb{1}(q_i\in s_j)\left(\log(p(s_j|q_i;\widetilde{\theta}))-\log\left(\frac{1}{N}\right)\right).
	\end{equation}
	Since each $q_i$ can only belong to one $s_j$ (i.e., one class), we can further simplify the objective function:
	\begin{equation}
	    \sum\limits_{i=1}^Q\sum\limits_{j=1}^N\mathbb{1}(q_i\in s_j)\log(p(s_j|q_i;\widetilde{\theta}))=\sum\limits_{i=1}^Q\log(p(s'_i|q_i;\widetilde{\theta})),
	\end{equation}
	where $s'_i$ denotes the specific $s_j$ that $q_i$ belongs to (i.e., $q_i\in s'_i$). Moreover, since $\log\left(1/N\right)$ is also a constant, we can simplify the objective function as follows:
	\begin{equation}
	\begin{aligned}
	    I(\mathbf{Q};\mathbf{S})
	    &=\sum\limits_{i=1}^Q\log(p(s'_i|q_i;\widetilde{\theta})).
	\end{aligned}
	\end{equation}
	To estimate $p(s'_i|q_i;\widetilde{\theta})$, we can define the probability of $q_i$ belonging to $s_j$ according to the squared $\ell_2$ norm of the embedding distance. Specifically, we further assign a weight parameter $\tau_i$ to each class and normalize the probability with a softmax function:
    \begin{equation}
        \left.p(s'_i|q_i;\widetilde{\theta})=\frac{\exp\left(-(\mathbf{q}_i- \mathbf{s}'_i)^2/\tau_i'\right)}{\sum_{j=1}^N\exp\left(-(\mathbf{q}_i-\mathbf{s}_j)^2/\tau_j\right)}
        \right.,
    \end{equation}
    where $\mathbf{q}_i$ and $\mathbf{s}_j$ denote the representations of the $i$-th query node $x_i$ in $\mathcal{Q}$ and the $j$-th class-ego subgraph in $\mathcal{S}$, respectively. $\tau_i$ is the adaptation parameter of the $i$-th class, and $\mathbf{s}'_i$ and $\tau_i'$ denote the specific class-ego subgraph representation and the adaptation parameter of the class that $\mathbf{q}_i$ belongs to, respectively. Then if we further apply the $\ell_2$ normalization to both $\mathbf{q}_i$ and $\mathbf{s}_i$, we obtain $(\mathbf{q}_i- \mathbf{s}_i)^2=2-2\mathbf{q}_i\cdot \mathbf{s}_i$. Combining the above equations, we can present the final optimization problem as follows:
    \begin{equation}
       \max_{\widetilde{\theta}} I(\mathcal{Q};\mathcal{S})=\min_{\widetilde{\theta}}\sum\limits_{i=1}^Q-\log\frac{\exp(\mathbf{q}_i\cdot \mathbf{s}'_i/\tau'_i)}{\sum_{j=1}^N\exp(\mathbf{q}_i\cdot \mathbf{s}_j/\tau_j)}.
       \label{eq:objective}
    \end{equation}
     %
     Since the distribution of node embeddings in each class $\mathcal{S}_i$ differs around the class representation $\mathbf{s}_i$, the node embeddings of specific classes can be scattered in a large distance from $\mathbf{s}_i$. In this case, the classification process should incorporate the distribution of node embeddings in each class while considering the entire support set $\mathcal{S}$. Thus, we propose to obtain $\tau_i'$ as follows:
	\begin{equation}
	    \tau_i=\frac{N\sum_k^K\|\mathbf{s}_i^k-\mathbf{s}_i\|_2}{\sum_j^N\sum_k^K\|\mathbf{s}_j^k-\mathbf{s}_j\|_2},
	\end{equation}
	where $\{\mathbf{s}_i^k\}_{i=1}^K$ denotes the node embeddings in the $i$-th class processed by $\text{GNN}_{\theta_i}$. In this way, the classification process is adapted regrading the entire support set to reduce the adverse impact of task variance. Hence, the model is able to incorporate information in the entire support set to achieve task-level adaptations. Then according to Eq.~(\ref{eq:objective}), we can optimize the following information loss to preserve the mutual information $I(\mathcal{Q};\mathcal{S})$:
	\begin{equation}
	    \mathcal{L}_{N}=-\sum\limits_{i=1}^Q\log\frac{\exp(\mathbf{q}_i\cdot \mathbf{s}'_i/\tau'_i)}{\sum_{j=1}^N\exp(\mathbf{q}_i\cdot \mathbf{s}_j/\tau_j)}.
	    \label{eq:Loss_N}
	\end{equation}
	
	It is worth mentioning that $\mathcal{L}_N$ shares a similar expression with the InfoNCE loss~\cite{oord2018representation,he2020momentum}. Thus, InfoNCE can be considered as a special case where positive and negative pairs are defined by different views of nodes, while in our case, they are defined according to the labels of query nodes. $\mathcal{L}_N$ also differs from the supervised contrastive loss~\cite{khosla2020supervised}, which utilizes various views of images and supervised information, as our loss aims at maximally preserving the mutual information between query nodes and support nodes within each meta-task. Moreover, the adaptation parameter acts similarly with the temperature parameter in InfoNCE, while in our framework, it is adjustable and provides task-level adaptations.

	\begin{algorithm}[t]
		\caption {Detailed learning process of our framework.}
		\begin{algorithmic}[1]
			\REQUIRE A graph $G=(\mathcal{V},\mathcal{E},\X)$, a meta-test task $\mathcal{T}_{test}=\{\mathcal{S},\mathcal{Q}\}$, base classes $\mathcal{C}_b$, meta-training epochs $T$, the number of classes $N$, and the number of labeled nodes for each class $K$.
			\ENSURE Predicted labels of the query nodes in $\mathcal{Q}$.
			
			// \texttt{Meta-training phase}
			\FOR {$i=1,2,\dotsc,T$}
			\STATE Sample a meta-training task $\mathcal{T}_{i}=\{\mathcal{S}_i,\mathcal{Q}_i\}$ from $\mathcal{C}_{b}$;
			\STATE Compute first-step node representations with $\text{GNN}_\phi$;
			\STATE Construct a class-ego subgraphs for each of $N$ classes in $\mathcal{S}_i$;
			\STATE Adapt $\text{GNN}_\theta$ to $\mathcal{T}_i$ according to Eq. (\ref{eq:adapt_class}) and (\ref{eq:adapt});
			\STATE Compute the representations for class-ego subgraphs and query nodes with the adapted $\text{GNN}_{\theta_i}$ and $\text{GNN}_{\theta_q}$;
			\STATE Update model parameters with the meta-training loss of $\mathcal{T}_i$ according to Eq. (\ref{eq:final_loss}) by one gradient descent step;
			\ENDFOR
			
			// \texttt{Meta-test phase}
			\STATE Compute first-step node representations with $\text{GNN}_\phi$;
			\STATE Construct a class-ego subgraphs for each of $N$ classes in $\mathcal{S}$;
			\STATE Adapt $\text{GNN}_\theta$ to $\mathcal{T}_{test}$ according to Eq. (\ref{eq:adapt_class}) and (\ref{eq:adapt});
			\STATE Compute the representations for class-ego subgraphs and query nodes with the adapted $\text{GNN}_{\theta_i}$ and $\text{GNN}_{\theta_q}$;
			\STATE Predict labels for query nodes in $\mathcal{Q}$;

		\end{algorithmic}
					\label{algorithm}
					
	\end{algorithm}

	\subsection{Few-shot Node Classification}
	So far, we can train $\text{GNN}_\phi$ and $\text{GNN}_\theta$ with the proposed loss $\mathcal{L}_N$ in Eq.~(\ref{eq:Loss_N}). However, since $\text{GNN}_\phi$ provides the first-step representations for nodes from an overview of the entire graph $G$, the supervised information within each meta-task could be insufficient for the optimization of $\text{GNN}_\phi$. Thus, we propose to classify query nodes from base classes $\mathcal{C}_{b}$ to optimize $\text{GNN}_\phi$. Specifically, we utilize an MLP layer followed by a softmax function to calculate the cross-entropy classification loss over $\mathcal{C}_{b}$:
	\begin{equation}
	    \mathbf{p}_i=\text{Softmax}\left(\text{MLP}(\mathbf{h}_i)   \right),
	\end{equation}
	\begin{equation}
	    \mathcal{L}_{CE}=-\sum\limits_{i=1}^Q\sum\limits_{j=1}^{|\mathcal{C}_b|}y_{i,j}\log p_{i,j},
	\end{equation}
	where $\mathbf{p}_i\in\mathbb{R}^{|\mathcal{C}_b|}$ is the probability that the $i$-th query node in $\mathcal{Q}$ belongs to each class in $\mathcal{C}_b$. $y_{i,j}=1$ if the $i$-th node belongs to the $j$-th class, and $y_{i,j}=0$, otherwise. $p_{i,j}$ is the $j$-th element in $\mathbf{p}_i$. In this way, instead of classifying nodes only from classes in a meta-task, we can utilize the supervised information in $\mathcal{C}_b$
	from a global perspective. Then the meta-training loss is defined as follows:
	\begin{equation}
	    \mathcal{L}=\mathcal{L}_{N}+\gamma\mathcal{L}_{CE},
	    \label{eq:final_loss}
	\end{equation}
	where $\gamma$ is an adjustable weight hyper-parameter.
	
	After meta-training, the meta-test process is the same as the meta-training process, except that meta-test tasks are sampled from novel classes $\mathcal{C}_n$. 
	The labels of query nodes are obtained by $\hat{y}_i=\argmax_j\{\mathbf{q}_i\cdot\mathbf{s}_j/\tau_i|j=1,2,\dotsc,N\}$, where $\mathbf{q}_i$ and $\mathbf{s}_j$ are representations of the $i$-th query node and the $j$-th class-ego subgraph, respectively. $\tau_i$ is the adaptation parameter of the $i$-th class. The detailed process of our framework is demonstrated in Algorithm \ref{algorithm}.

    \section{Experiments}
    In this section, we conduct experiments to evaluate our framework \textsc{TENT} on four prevalent few-shot node classification datasets. Furthermore, we conduct experiments to verify the effectiveness of different modules in our framework with ablation study and demonstrate the parameter sensitivity. 

\subsection{Datasets}
To evaluate our framework on few-shot node classification tasks, we conduct experiments on four prevalent real-world graph datasets: \texttt{Amazon-E}~\cite{mcauley2015inferring}, \texttt{DBLP}~\cite{tang2008arnetminer}, \texttt{Cora-full}~\cite{bojchevski2018deep}, and \texttt{OGBN-arxiv}~\cite{hu2020open}. 
We summarize the detailed statistics of these datasets in Table~\ref{tab:statistics}. Specifically,
\# Nodes and \# Edges denote the number of nodes and edges in the graph, respectively. \# Features denotes the dimension of node features. 
Class Split denotes the number of classes used for meta-training/validation/meta-test. More details are provided in Appendix \ref{appendix}.

	\begin{table}[htbp]
	
		\setlength\tabcolsep{6pt}
		\small
		\centering
		\renewcommand{\arraystretch}{1.2}
		\caption{Statistics of four node classification datasets. }
        \vspace{-0.1in}
		\begin{tabular}{c|c|c|c|c}
		\hline
        \textbf{Dataset}&\# Nodes & \# Edges & \# Features & Class Split\\
        \hline
        \texttt{Amazon-E}&42,318& 43,556&8,669& 90/37/40\\
        \texttt{DBLP}&40,672&288,270&7,202&80/27/30\\
        \texttt{Cora-full}&19,793&65,311&8,710&25/20/25\\
        \texttt{OGBN-arxiv}&169,343&1,166,243&128&15/5/20\\
        
        \hline
		\end{tabular}
        \vspace{-0.1in}
		\label{tab:statistics}
	\end{table}

\subsection{Experimental Settings}
To validate the effectiveness of our proposed framework \textsc{TENT}, we conduct experiments with the following baseline methods to compare performance:
\begin{itemize}
    \item \textbf{Prototypical Networks}~\cite{snell2017prototypical}: Prototypical Networks learn prototypes for classes for query matching.
        \item \textbf{MAML}~\cite{finn2017model}: MAML proposes to optimize model parameters based on gradients of support instances across meta-tasks.
    \item \textbf{GCN}~\cite{kipf2017semi}: GCN performs information propagation based on local structures.

        \item \textbf{G-Meta}~\cite{huang2020graph}: G-Meta utilizes representations of subgraphs as node embeddings for few-shot learning on graphs.
    \item \textbf{GPN}~\cite{ding2020graph}: GPN leverages node importance and Prototypical Networks to improve performance.

    \item \textbf{RALE}~\cite{liu2021relative}: RALE proposes to learn node dependencies according to node locations on the graph.
\end{itemize}

During training, we sample a certain number of meta-training tasks from training classes (i.e., base classes) and train the model with these meta-tasks. Then we evaluate the model based on a series of randomly sampled meta-test tasks from test classes (i.e., novel classes). For consistency, the class splitting is identical for all baseline methods. Then the final result of the average classification accuracy is obtained based on these meta-test tasks. More detailed parameter settings can be found in Appendix~\ref{appendix:implementation}.

	\begin{table*}[htbp]
		\small
		\centering
		\renewcommand{\arraystretch}{1.2}
		\caption{The overall few-shot node classification results (accuracy in \%) of various models under different few-shot settings.}

		\begin{tabular}{c||c|c|c|c||c|c|c|c}
			\hline
			Dataset&\multicolumn{4}{c||}{\texttt{DBLP}}&\multicolumn{4}{c}{\texttt{Amazon-E}}
			\\
			\hline
			Setting&5-way 3-shot&5-way 5-shot&10-way 3-shot& 10-way 5-shot&5-way 3-shot&5-way 5-shot&10-way 3-shot& 10-way 5-shot\\
			\hline\hline
			PN~\cite{snell2017prototypical}&$41.51\pm3.60$&$46.17\pm3.55$&$28.98\pm3.87$&$36.71\pm3.35$&$56.80\pm3.60$&$62.53\pm2.80$&$44.26\pm2.64$&$48.20\pm3.89$\\\hline
			MAML~\cite{finn2017model}&$43.06\pm2.92$&$49.93\pm2.57$&$34.63\pm3.91$&$38.44\pm3.25$&$56.03\pm2.11$&$63.40\pm3.33$&$40.80\pm2.75$&$47.06\pm3.15$\\\hline
			GCN~\cite{kipf2017semi}&$62.87\pm1.44$&$70.51\pm1.37$&$47.22\pm2.97$&$53.95\pm2.49$&$55.33\pm1.23$&$62.96\pm2.61$&$45.18\pm2.61$&$50.89\pm2.95$\\\hline
			G-Meta~\cite{huang2020graph}&$73.49\pm2.82$&$78.56\pm2.86$&$60.77\pm3.03$&$66.26\pm3.47$&$64.56\pm4.23$&$68.36\pm4.10$&$59.75\pm4.90$&$63.02\pm4.11$\\\hline
            GPN~\cite{ding2020graph}&$76.42\pm3.11$&$80.85\pm3.68$&$63.14\pm2.25$&$69.55\pm2.56$&$65.16\pm3.17$&$71.89\pm3.94$&$62.52\pm3.12$&$63.98\pm2.04$\\\hline
            RALE~\cite{liu2021relative}&$75.38\pm4.94$&$79.85\pm4.69$&$62.81\pm3.48$&$67.61\pm3.99$&$69.55\pm4.24$&$74.97\pm4.66$&$63.27\pm3.31$&$64.85\pm3.04$\\\hline
 TENT&\textbf{79.04} $\pm$ \textbf{3.14}&\textbf{82.84} $\pm$ \textbf{3.97}&\textbf{65.47} $\pm$ \textbf{4.21}&\textbf{72.38} $\pm$ \textbf{4.14}&\textbf{75.76} $\pm$ \textbf{3.63}&\textbf{79.38} $\pm$ \textbf{4.98}&\textbf{67.59} $\pm$ \textbf{4.16}&\textbf{69.77} $\pm$ \textbf{3.76}

\\\hline

		\end{tabular}
		\label{tab:all_result}
		\vspace{-0.1cm}
	\end{table*}
	
	\begin{table*}[htbp]
		\small
		\centering
		\renewcommand{\arraystretch}{1.2}

		\begin{tabular}{c||c|c|c|c||c|c|c|c}
			\hline
			Dataset&\multicolumn{4}{c||}{\texttt{Cora-full}}&\multicolumn{4}{c}{\texttt{OGBN-arxiv}}
			\\
			\hline
			Setting&5-way 3-shot&5-way 5-shot&10-way 3-shot& 10-way 5-shot&5-way 3-shot&5-way 5-shot&10-way 3-shot& 10-way 5-shot\\
			\hline\hline
			PN~\cite{snell2017prototypical}&$42.62\pm3.78$&$56.66\pm2.91$&$35.95\pm3.95$&$38.69\pm3.09$&$37.99\pm3.98$&$49.71\pm4.20$&$31.44\pm3.00$&$35.79\pm3.63$\\\hline
			MAML~\cite{finn2017model}&$47.10\pm4.32$&$54.89\pm3.09$&$30.68\pm3.08$&$42.22\pm2.76$&$41.83\pm2.54$&$42.14\pm3.86$&$33.15\pm2.92$&$36.82\pm3.03$\\\hline
			GCN~\cite{kipf2017semi}&$49.05\pm2.04$&$58.03\pm3.50$&$34.27\pm3.98$&$39.85\pm3.50$&$44.80\pm2.56$&$47.29\pm3.58$&$35.80\pm2.21$&$37.78\pm2.90$\\\hline
			G-Meta~\cite{huang2020graph}&$57.93\pm3.79$&$60.30\pm2.93$&$45.67\pm3.35$&$47.76\pm3.25$&$47.66\pm3.27$&$49.81\pm4.01$&$35.93\pm3.04$&$40.13\pm4.35$\\\hline
            GPN~\cite{ding2020graph}&$58.38\pm3.49$&$63.82\pm2.93$&$41.65\pm2.20$&$45.63\pm3.17$&$49.16\pm3.43$&$53.06\pm3.13$&$37.28\pm3.99$&$43.33\pm3.27$\\\hline

            RALE~\cite{liu2021relative}&$62.83\pm3.12$&$65.93\pm3.24$&$48.05\pm3.09$&$51.67\pm3.21$&$53.90\pm3.45$&$56.99\pm4.43$&$37.60\pm4.12$&$41.42\pm3.03$\\\hline
            TENT&\textbf{64.80} $\pm$ \textbf{4.10}&\textbf{69.24} $\pm$ \textbf{4.49}&\textbf{51.73} $\pm$ \textbf{4.34}&\textbf{56.00} $\pm$ \textbf{3.53}&\textbf{55.62} $\pm$ \textbf{3.13}&\textbf{62.96} $\pm$ \textbf{3.74}&\textbf{41.13} $\pm$ \textbf{4.26}&\textbf{44.73} $\pm$ \textbf{3.42}
\\\hline

		\end{tabular}

	\end{table*}

\subsection{Overall Evaluation Results}
We first present the performance comparison of our framework and baseline methods on few-shot node classification in Table~\ref{tab:all_result}. Specifically, to better demonstrate the efficacy of our framework under different few-shot settings, we conduct the experiments under four different settings: 5-way 3-shot, 5-way 5-shot, 10-way 3-shot, and 10-way 5-shot. Moreover, the evaluation metric is the average classification accuracy over ten repetitions. From the overall results, we can obtain the following observations:
\begin{itemize}
    \item Our proposed framework TENT outperforms all other baselines in all datasets under different few-shot settings, which validates the effectiveness of our task-adaptive framework on few-shot node classification.
    \item Conventional few-shot methods such as Prototypical Network~\cite{snell2017prototypical} and MAML~\cite{finn2017model} exhibit inferior performance compared with other baselines. The reason is that such methods are proposed in other domains and thus result in unsatisfactory performance on graphs.
    \item When increasing the value of $K$ (i.e., more support nodes in each class), all methods gain considerable performance improvements. Moreover, our framework achieves better results due to that the node-level and class-level adaptations benefit more from a larger size of nodes in each class.
    \item The performance of all methods significantly decreases when the value of $N$ increases (i.e., more classes in each meta-task). The main reason is that the variety of classes in each meta-task leads to a more complex class distribution and results in classification difficulties. However, by incorporating the class-level and task-level adaptations, our framework is capable of alleviating this problem 
    when a larger $N$ is presented.
\end{itemize}

		\begin{figure}[!t]
		\centering
\captionsetup[sub]{skip=-1pt}
\subcaptionbox{\texttt{DBLP}}
{\includegraphics[width=0.23\textwidth]{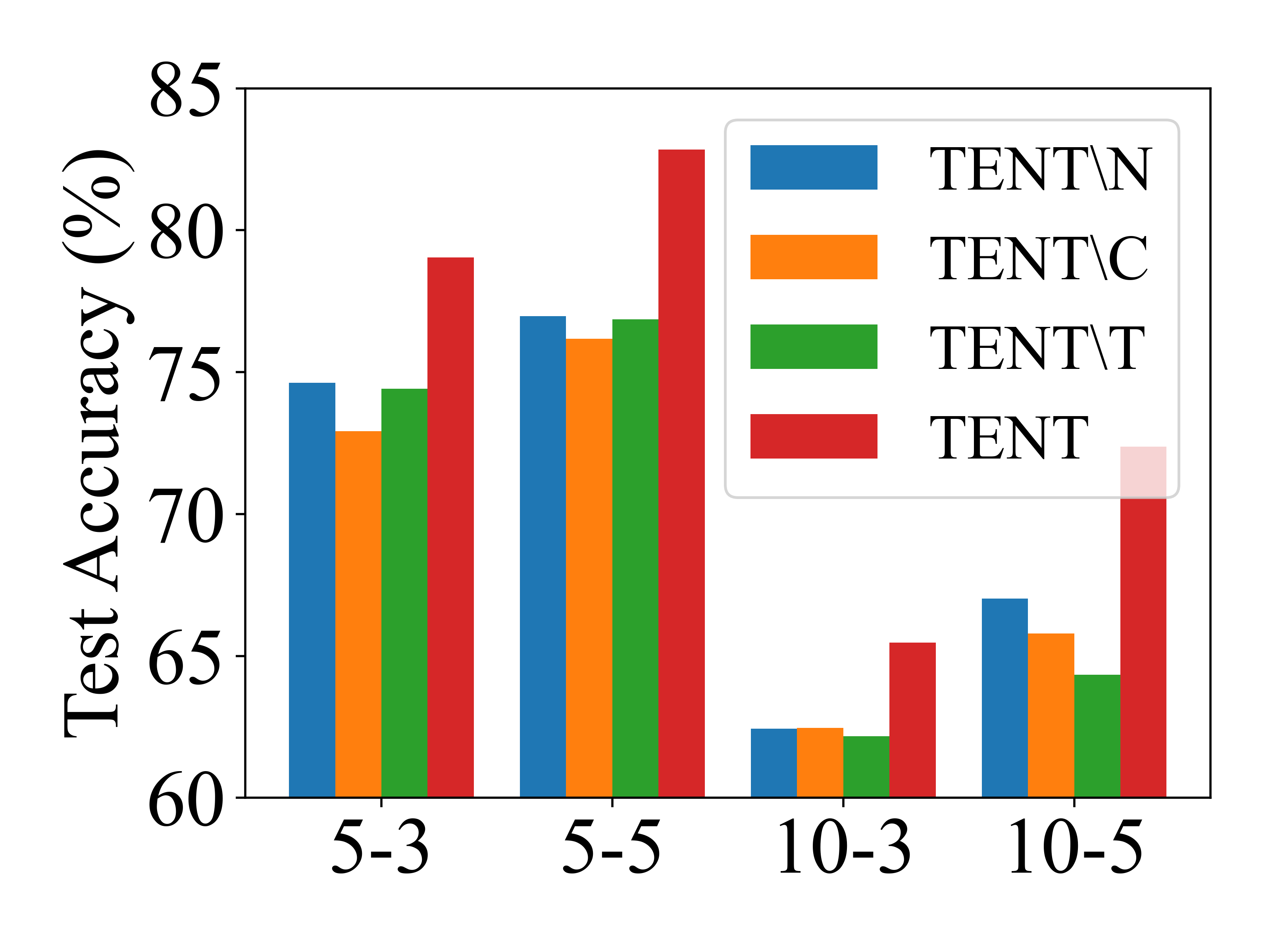}}
\subcaptionbox{\texttt{Amazon-E}}
{\includegraphics[width=0.23\textwidth]{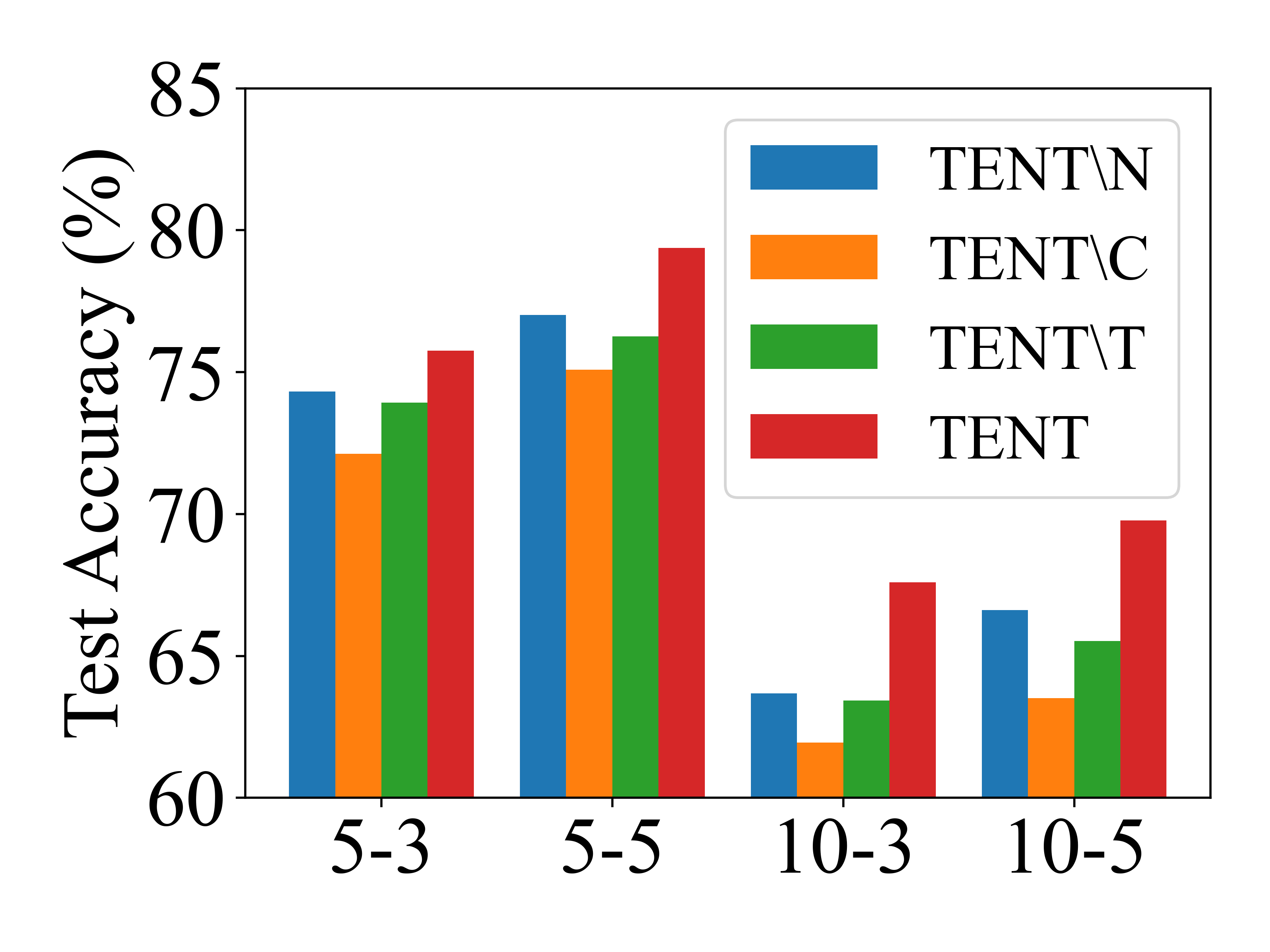}}
\subcaptionbox{\texttt{Cora-full}}
{\includegraphics[width=0.23\textwidth]{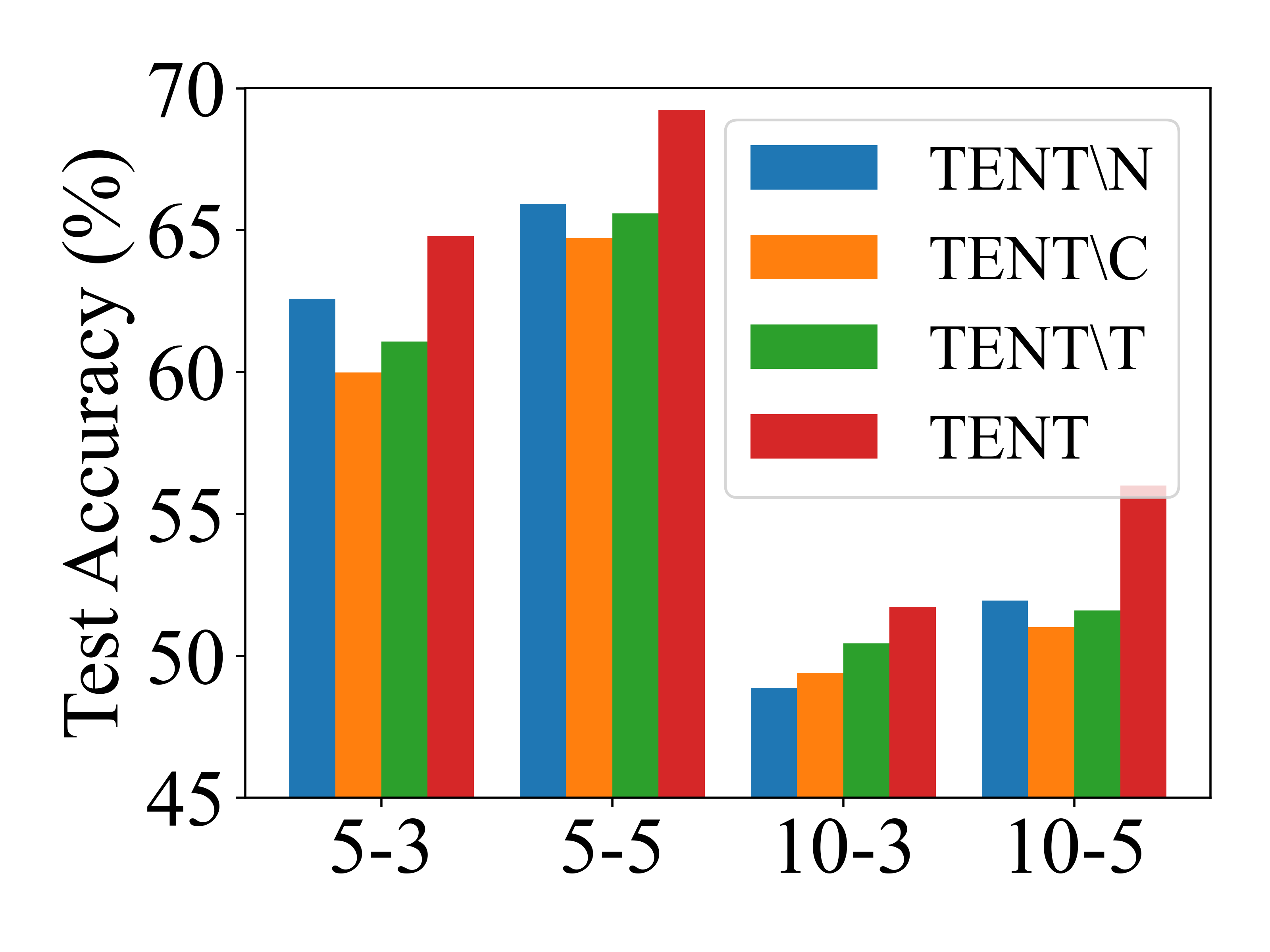}}
\subcaptionbox{\texttt{OGBN-arxiv}}
{\includegraphics[width=0.23\textwidth]{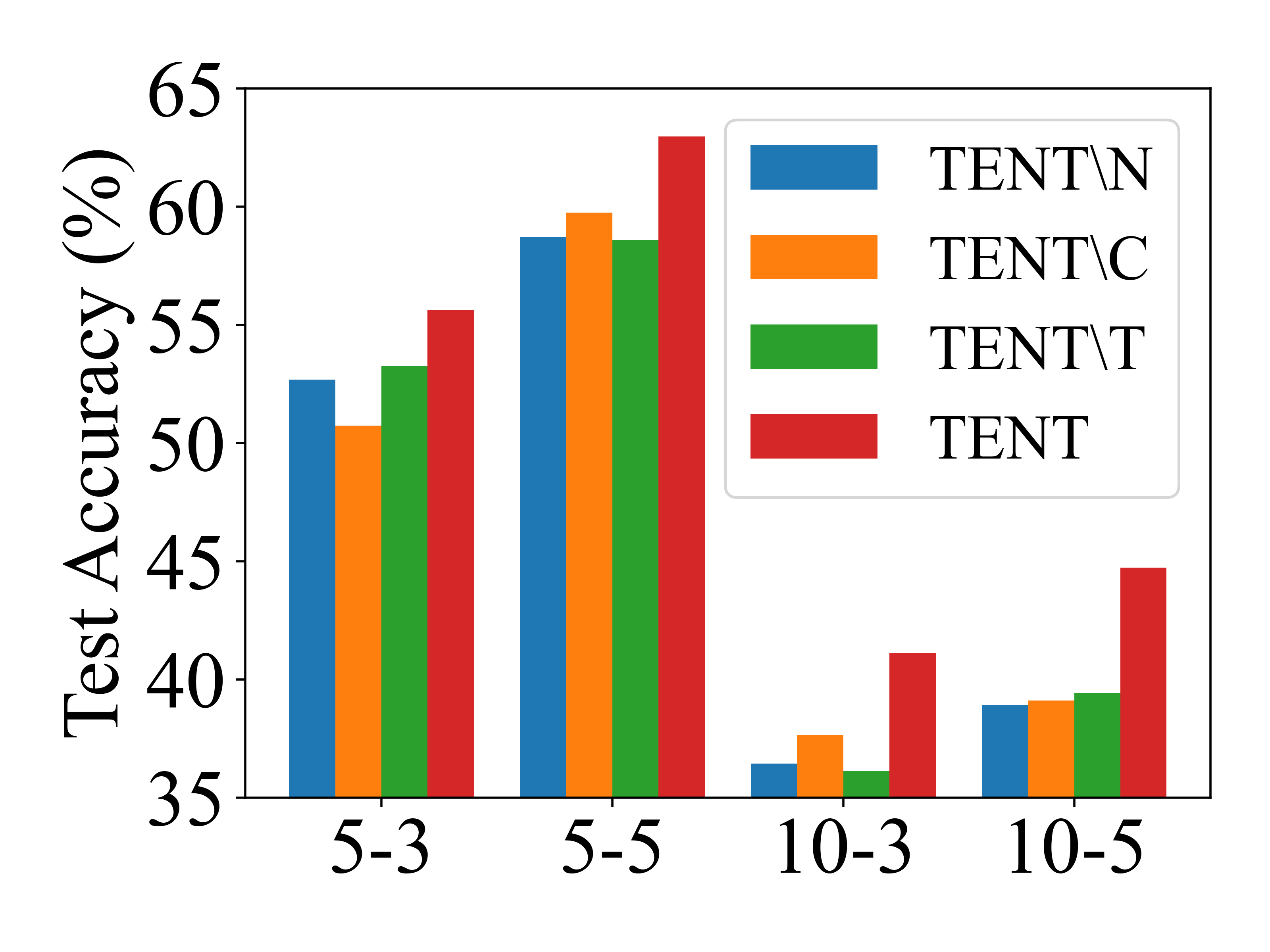}}
\vspace{-0.1in}
		\caption{Ablation study on our framework in the $N$-way $K$-shot setting.}
		\label{fig:ablation}
\vspace{-0.12in}
	\end{figure}

\subsection{Ablation Study}
In this part, we conduct an ablation study on four datasets to verify the importance of three crucial components in TENT. First, we remove the node-level adaptation and directly utilize the original graph instead of class-ego subgraphs to learn representations for each class in meta-tasks. In this way, the support nodes of different classes are distributed over the entire graph and thus lack node-level adaptations. We refer to this variant as \emph{TENT\textbackslash N}. Second, we remove the class-specific adapter so that the framework identically learns class representations and lacks class-level adaptations, and we refer to this variant as \emph{TENT\textbackslash C}. The final variant is to replace the task-level adaptation module with a common Euclidean distance classifier, which means during training, the framework fails to learn task-level adaptations across meta-training tasks, and we refer it to as \emph{TENT\textbackslash T}. The overall ablation study results are presented in Fig.~\ref{fig:ablation}. From the results, we can observe that 
TENT outperforms all variants, which demonstrates the effectiveness of all three types of adaptations. Specifically, removing node-level adaptations results in a large decrease in few-shot node classification performance. Furthermore, integrating class-level adaptations provides a considerable performance improvement, especially when the number of classes increases, which introduces larger class variance. More significantly, without the task-level adaptations, the performance decreases rapidly when the support set size increases. Therefore, the result further demonstrates the importance of task-level adaptations in the presence of a more complex few-shot setting with a large support set.

		\begin{figure}[!t]
		\centering
\captionsetup[sub]{skip=-2pt}
\subcaptionbox{\texttt{DBLP}}
{\includegraphics[width=0.23\textwidth]{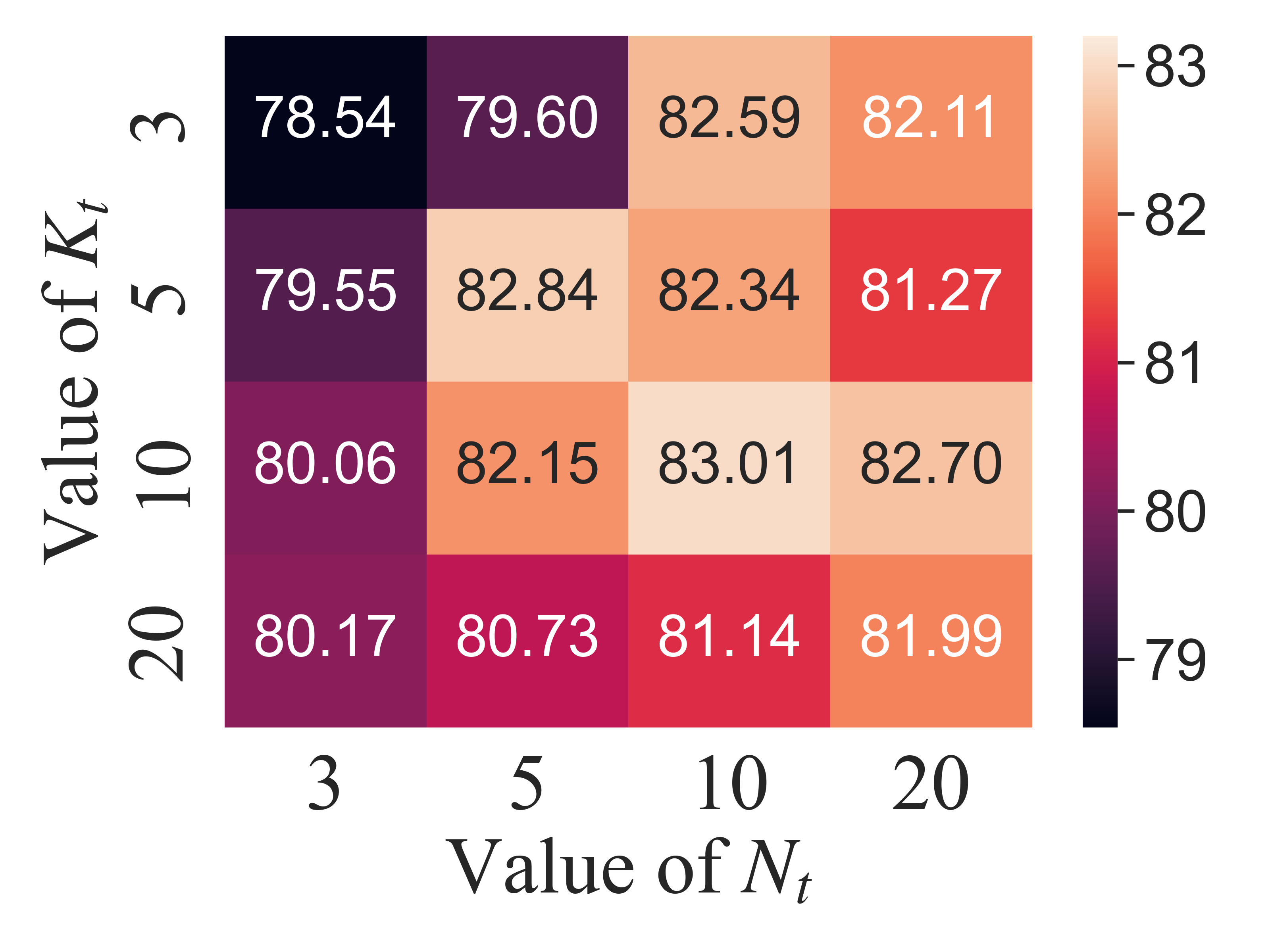}}
\subcaptionbox{\texttt{Amazon-E}}
{\includegraphics[width=0.23\textwidth]{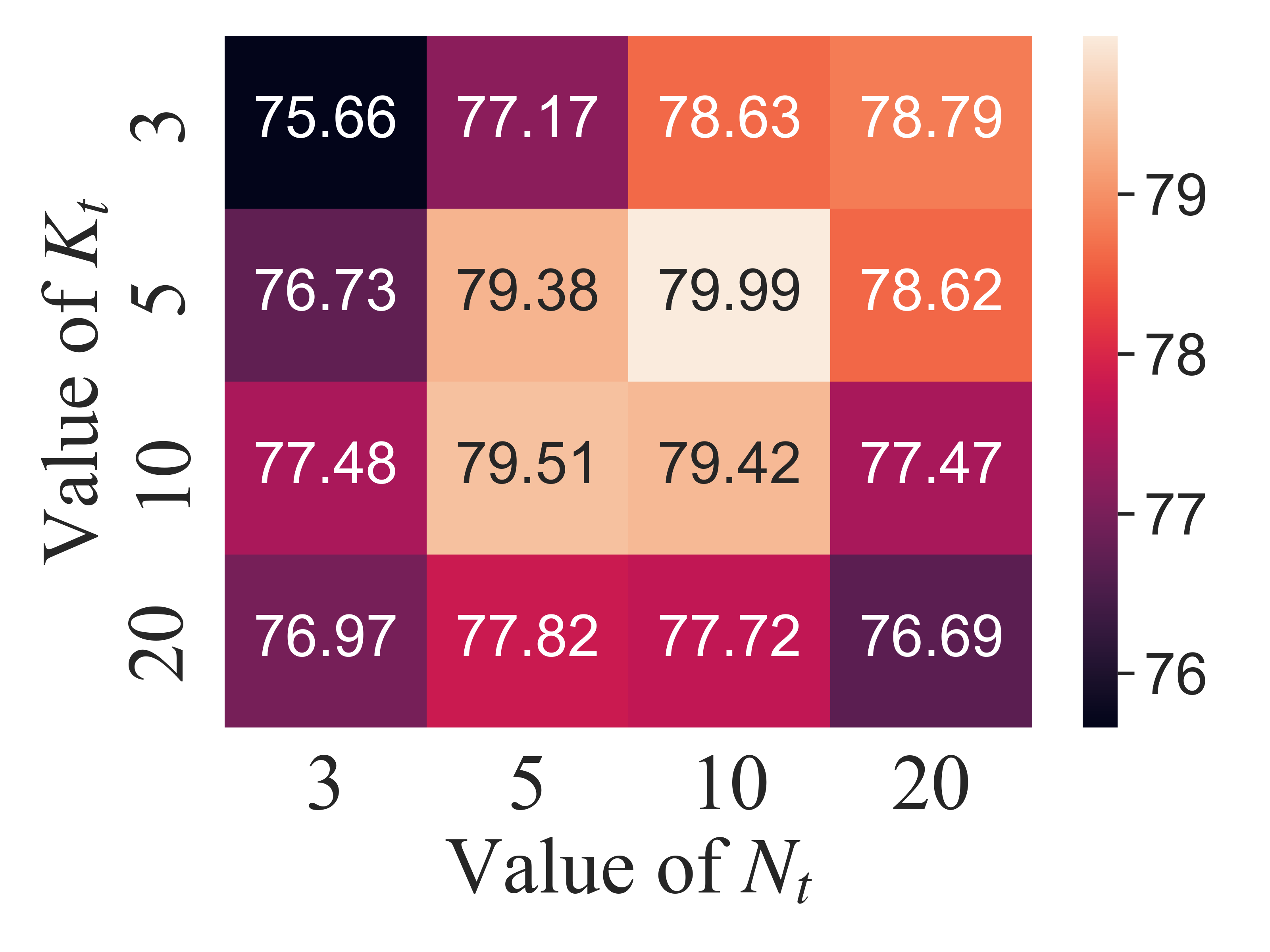}}
\subcaptionbox{\texttt{Cora-full}}
{\includegraphics[width=0.23\textwidth]{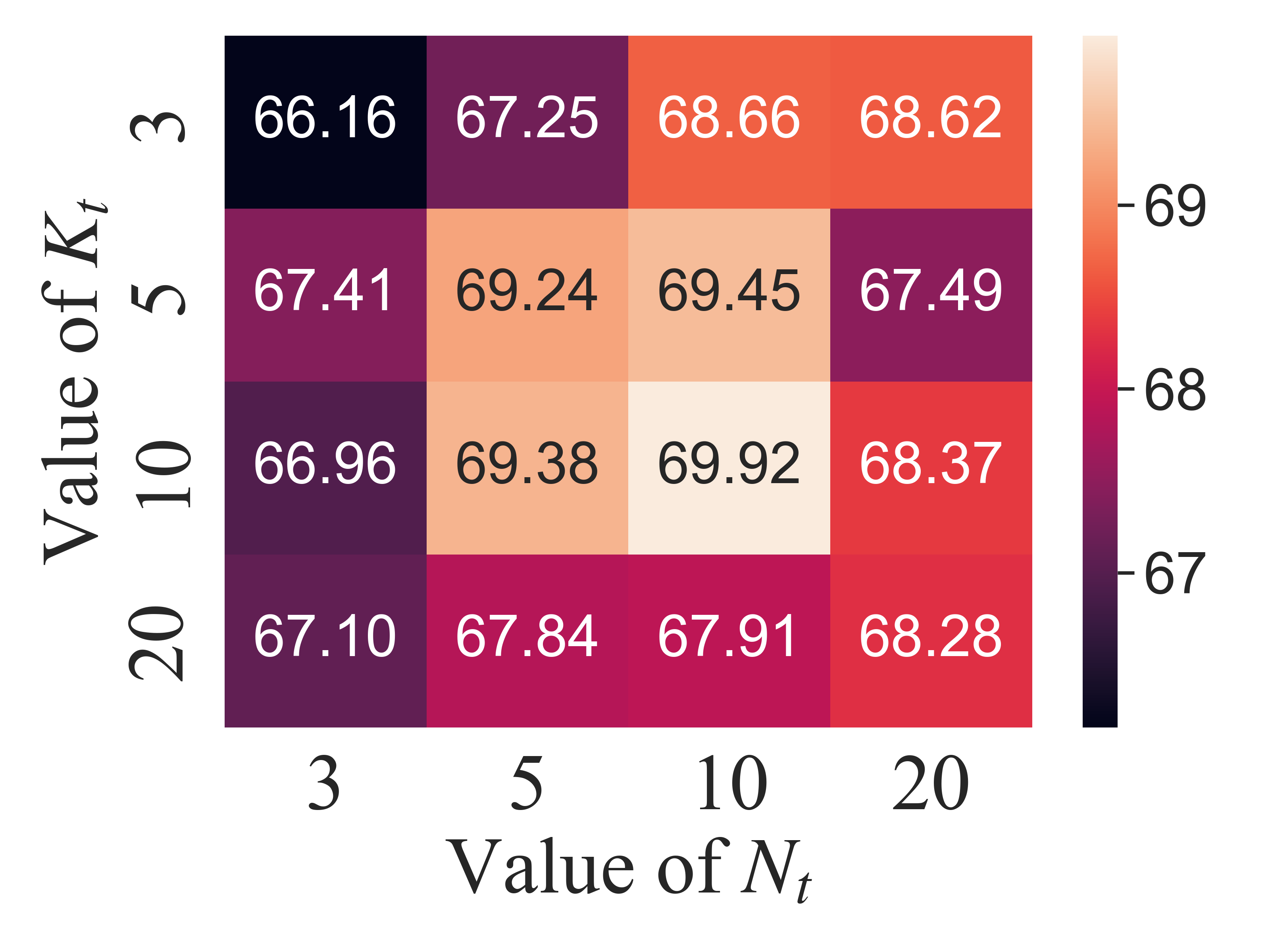}}
\subcaptionbox{\texttt{OGBN-arxiv}}
{\includegraphics[width=0.23\textwidth]{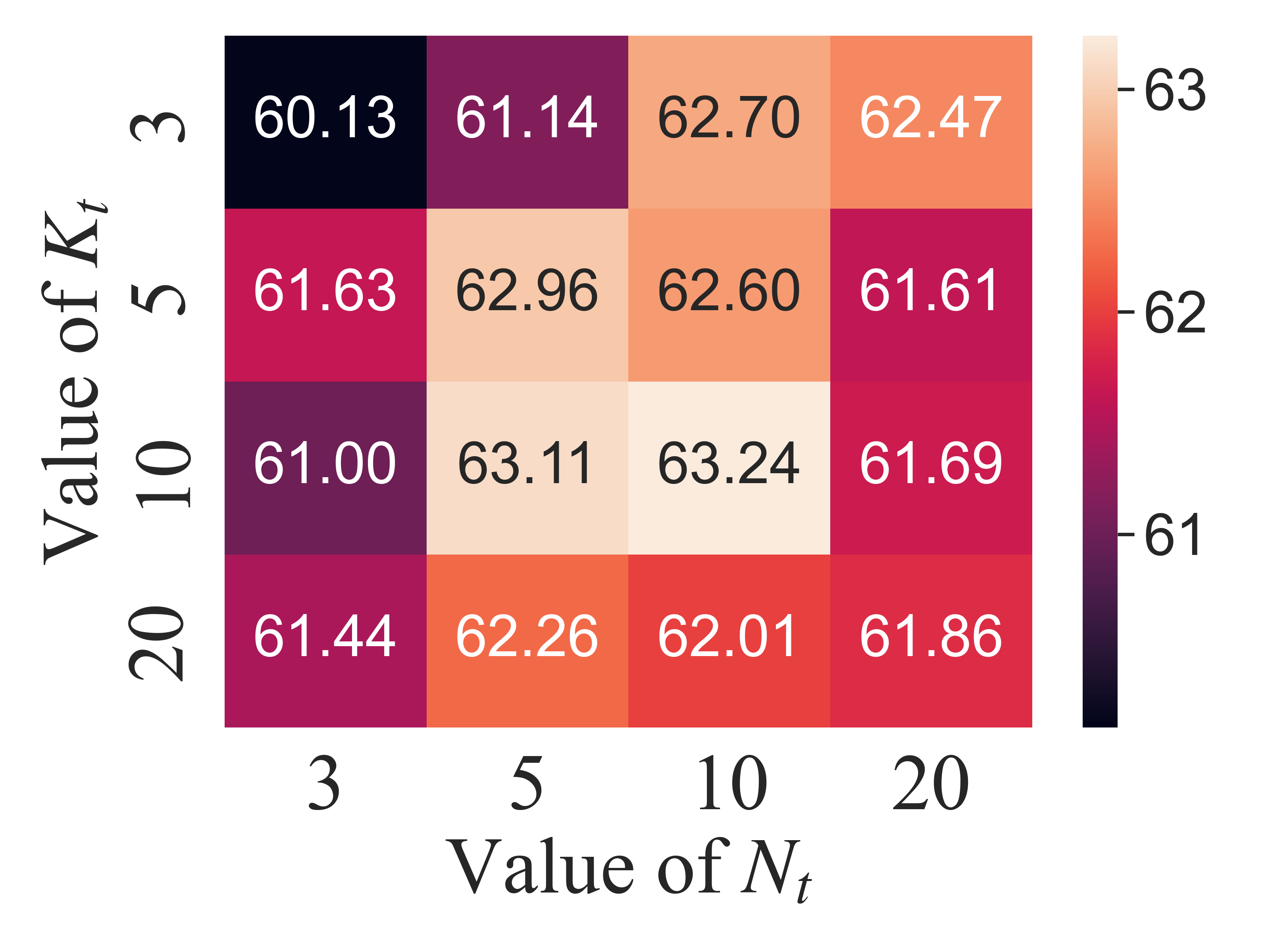}}
\vspace{-0.1in}
		\caption{Results of TENT with different $N_t$ and $K_t$.}
		\label{fig:nk}
\vspace{-0.12in}
	\end{figure}

\subsection{Effect of Meta-training Support Set Size}
In this section, we conduct experiments to study the sensitivity of several parameters in TENT. Since TENT provides task adaptations for both meta-training and meta-test tasks, the values of $N$ (i.e., number of classes in a support set) and $K$ (i.e., number of support nodes in each class) are unnecessary to be consistent during meta-training and meta-test. In other words, it differs from the general few-shot learning setting, where the parameters of $N$ and $K$ are consistent during meta-training and meta-test. Therefore, we can adjust these two parameters during meta-training to analyze their effects for better performance. Fig.~\ref{fig:nk} reports the classification accuracy of TENT when varying the parameters of $N$ and $K$ during meta-training on four datasets, denoted as $N_t$ and $K_t$, respectively. Specifically, we vary the values of $N_t$ and $K_t$ as 3, 5, 10, and 20. Note that during meta-test, the values of $N$ and $K$ are kept invariant as 5 and 5, respectively (i.e., $5$-way $5$-shot). From the results, we observe that increasing $N_t$ and $K_t$ both provide better results on few-shot node classification. The reason is that TENT learns the three types of adaptations from a larger support set during meta-training and thus is more capable of handling 
node-level and class-level variance. More specifically, increasing the value of $N$ results in a more significant improvement. The main reason is that the class-level and task-level adaptations benefit from more classes in each meta-task. In addition, incorporating more support nodes in each class (i.e., larger $K_t$) also enhances the interactions 
among nodes in each class-ego subgraph for more comprehensive node-level adaptations.

		\begin{figure}[!t]
		\centering
\captionsetup[sub]{skip=-2pt}
\subcaptionbox{\texttt{DBLP}}
{\includegraphics[width=0.23\textwidth]{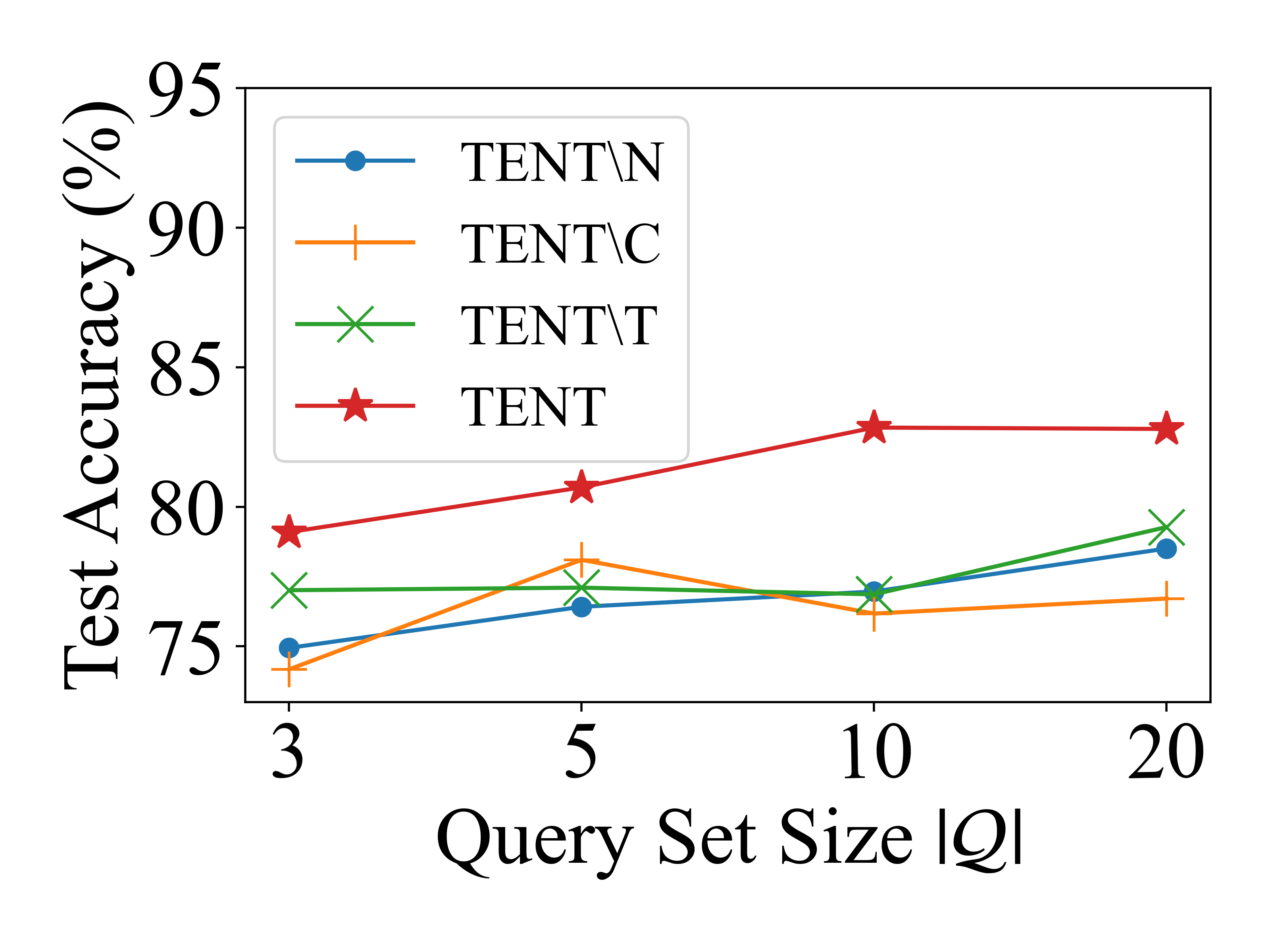}}
\subcaptionbox{\texttt{Amazon-E}}
{\includegraphics[width=0.23\textwidth]{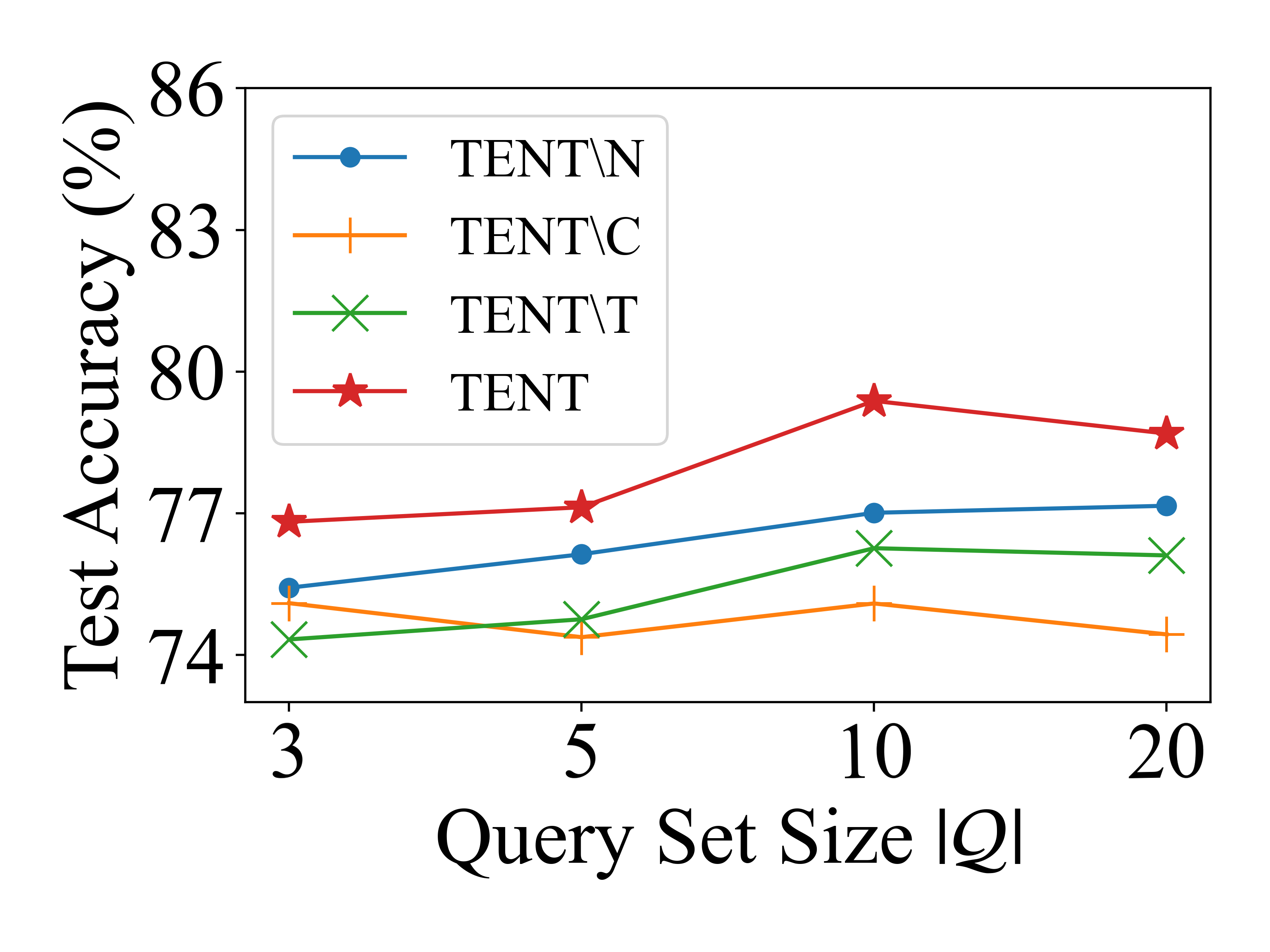}}
\subcaptionbox{\texttt{Cora-full}}
{\includegraphics[width=0.23\textwidth]{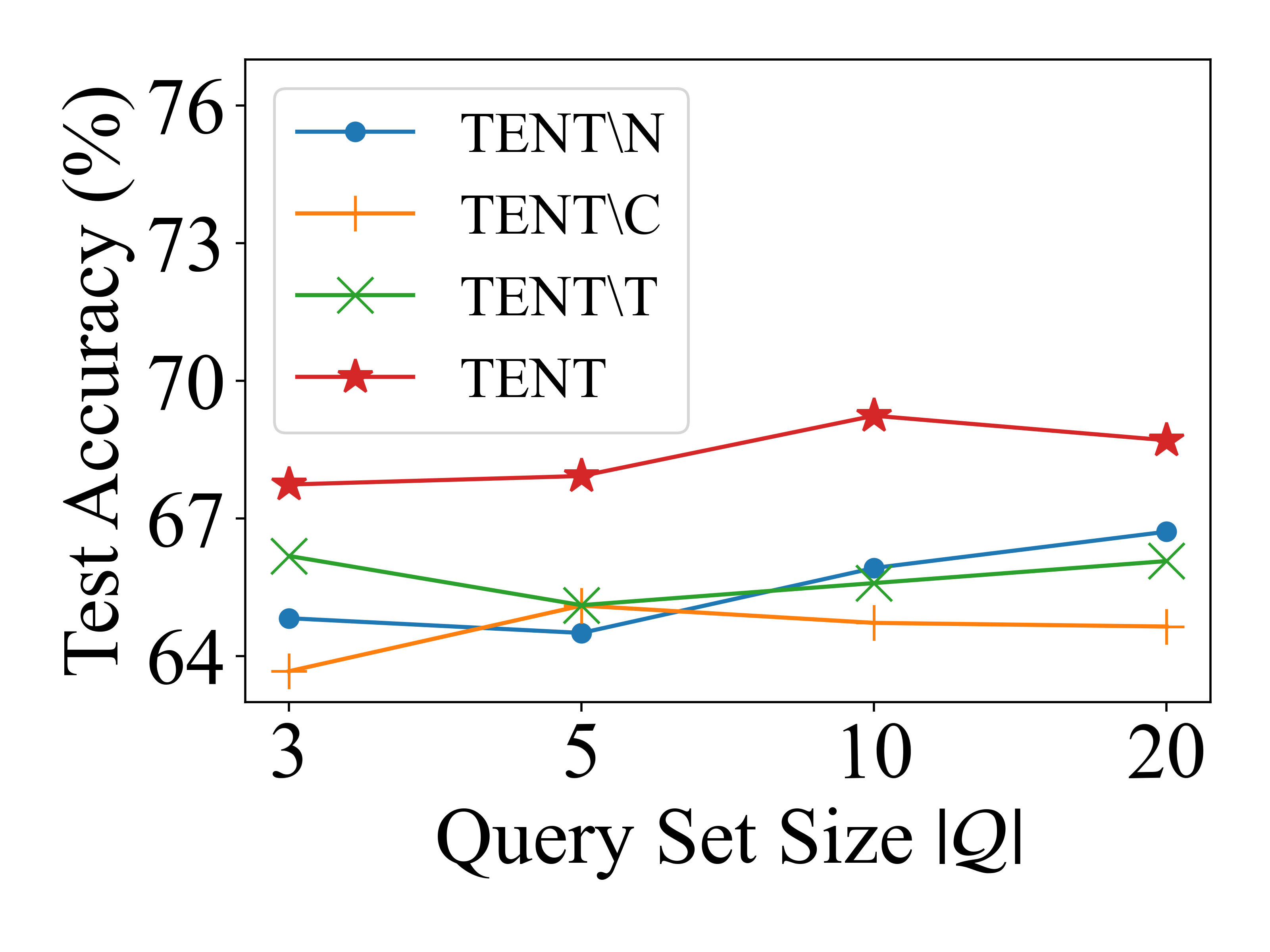}}
\subcaptionbox{\texttt{OGBN-arxiv}}
{\includegraphics[width=0.23\textwidth]{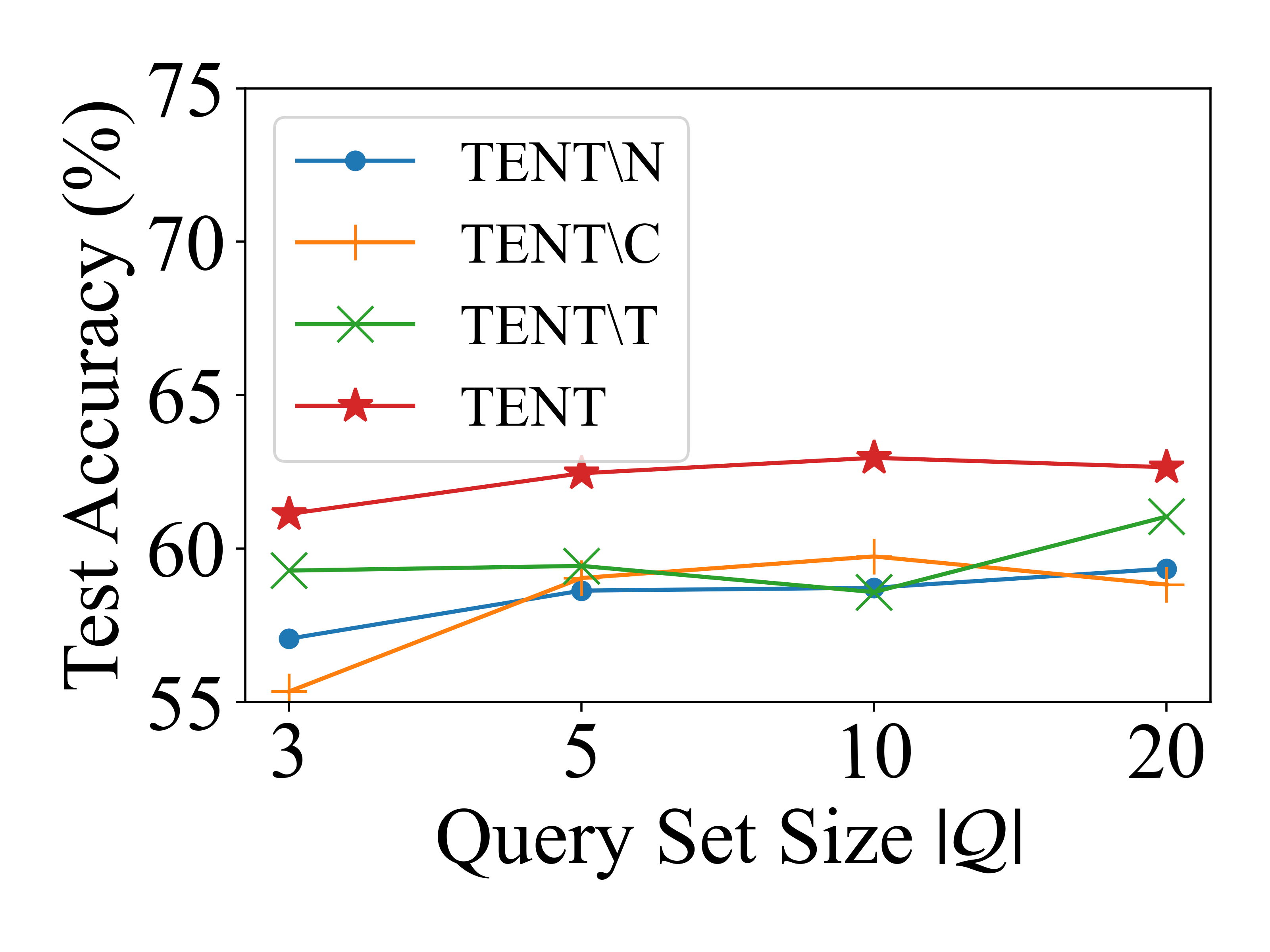}}

		\caption{Results of TENT with different values of $|\mathcal{Q}|$.}
		\label{fig:Q}
\vspace{-0.15in}
	\end{figure}

\subsection{Effect of Query Set Size}
In this part, we conduct experiments to present how the query set size $|\mathcal{Q}|$ in each meta-task during meta-training affects the performance of our proposed framework TENT. Fig.~\ref{fig:Q} reports the results of TENT when varying the value of $|\mathcal{Q}|$ on four datasets under the 5-way 5-shot setting. Specifically, $|\mathcal{Q}|$ during meta-training is changed from 3 to 20, while it remains 10 during meta-test for a fair comparison. From the results, we can observe that the few-shot node classification performance increases when $|\mathcal{Q}|$ becomes larger. The reason is mainly attributed to the fact that involving more
query nodes during meta-training (i.e., increasing the value of $|\mathcal{Q}|$) helps
alleviate the over-fitting problem. However, as the results suggest, an excessively large query set size may result in a performance drop. The reason is that the optimization process may be more difficult on a large query set.

\section{Related Work}
\subsection{Graph Neural Networks}
Recently, many researchers 
focus on studying Graph Neural Networks (GNNs) to learn comprehensive node representations in graphs~\cite{cao2016deep,chang2015heterogeneous,xu2018representation}. In general, GNNs aim at learning node representations through a certain number of information propagation steps in a recurrent manner~\cite{zhou2020graph,zhang2020deep,hamilton2017inductive}. 
In this way, GNNs can aggregate information from neighboring nodes to generate node representations based on local structures.
For example, Graph Convolutional Networks (GCNs)~\cite{kipf2017semi} perform convolution operations on graphs based on the graph spectral theory. 
Graph Attention Networks (GATs)~\cite{velivckovic2017graph} leverage the attention mechanism to select more important neighboring nodes for aggregation. Moreover, Graph Isomorphism Networks (GINs)~\cite{xu2018powerful} develop an expressive architecture, which is as powerful as the Weisfeiler-Lehman graph isomorphism test. 
Nonetheless, GNNs typically render sub-optimal performance when there are limited labeled nodes for each class~\cite{ding2020graph,yao2020graph},
which further indicates the necessity of few-shot learning on graphs.

\subsection{Few-shot Learning on Graphs}
Few-shot Learning (FSL) aims to learn transferable knowledge from tasks with abundant supervised information and generalize it to novel tasks with a limited number of labeled instances. In general, few-shot learning methods can be divided into two categories: \emph{metric-based} approaches and \emph{meta-optimizer-based} approaches. Specifically, the metric-based approaches aim at learning generalizable metric functions to match the query set with the support set for classification~\cite{liu2019learning,sung2018learning}. 
For example, Matching Networks~\cite{vinyals2016matching} conduct predictions based on the similarity between a query instance and each support instance learned by attention networks. Prototypical Networks~\cite{snell2017prototypical} learn a prototype as the representation for each class and perform classification based on the Euclidean distances between query instances and prototypes. 
On the other hand, meta-optimizer-based approaches aim at optimizing model parameters according to gradients calculated from few-shot instances~\cite{mishra2018simple,ravi2016optimization}. For example, MAML~\cite{finn2017model} optimizes model parameters based on gradients on support instances for fast generalization. Moreover, LSTM-based meta-learner~\cite{ravi2016optimization} proposes to adjust the step size for updating parameters during meta-training. 

In the field of graphs, several recent works propose to conduct graph-based tasks under the few-shot learning scenario~\cite{chauhan2020few,ma2020adaptive,wang2021reform}. Among them, 
GPN~\cite{ding2020graph} proposes to leverage node importance based on Prototypical Networks~\cite{snell2017prototypical} for better performance, where nodes are classified via finding the nearest class prototype. G-Meta~\cite{huang2020graph} leverages local subgraphs to learn node representations while combining meta-learning~\cite{finn2017model} for model generalization. More recently, RALE~\cite{liu2021relative} learns to model node dependencies within each meta-task by assigning relative and absolute locations for nodes with task-level and graph-level dependencies, respectively.

\section{Conclusion}
In this paper, we study the problem of few-shot node classification, which 
aims at predicting labels for nodes in novel classes with limited labeled nodes. Furthermore, to address the associated challenges caused by insufficient labeled nodes and the variety of novel classes, we propose a novel framework TENT to perform task adaptations for each meta-task from three perspectives: node-level, class-level, and task-level. As a result, our framework can perform these adaptions to each meta-task and advance classification performance with respect to a variety of novel classes during meta-test. Moreover, extensive experiments are conducted on four prevalent few-shot node classification datasets. The experimental results further validate that TENT outperforms other state-of-the-art baselines. In addition, the ablation study also verifies the effectiveness of three different levels of adaptations in our framework. Nevertheless, there still exists a considerable number of difficulties in few-shot node classification. For example, the inductive setting for few-shot node classification is still challenging. Future work may incorporate more sophisticated adaptation methods to handle the novel classes on graphs unseen during meta-training.

\section*{Acknowledgement}
This material is supported by the National Science Foundation (NSF) under grant \#2006844.

	\bibliographystyle{ACM-Reference-Format}

	\bibliography{acmart}

	\clearpage
	\begin{appendices}
	
	\section{Appendix}
	\subsection{Notations}
To provide better understandings, we present the
utilized notations in this paper and the corresponding descriptions.

\begin{table}[htbp]
\small
\setlength\tabcolsep{0.5pt}
\caption{Notations used in this paper.} 
\vspace{-0.2cm}
\label{tb:symbols}
\begin{tabular}{cc}

\hline

\textbf{Notations}       & \textbf{Definitions or Descriptions} \\
\hline

$G$   &  the input graph\\
$\mathcal{V}$, $\mathcal{E}$  & the node set and the edge set of $G$\\
$\X$ & the input node features of $G$\\
$\mathcal{C}_b$,$\mathcal{C}_n$ & the base class set and the novel class set\\
$\mathcal{T}_i$, $\mathcal{S}_i$, $\mathcal{Q}_i$& the $i$-th meta-task and its support set and query set\\
$\alpha_i$, $\beta_i$, $\tau_i$ & adaptation parameters for the $i$-th class\\
$N$&the number of support classes in each meta-task\\
$K$&the number of labeled nodes in each class\\
$N_t$, $K_t$& the value of $N$ and $K$ during meta-training\\
$\mathbf{s}_i$&the embedding of the $i$-th class in each meta-task\\
$\mathbf{q}_i$&the embedding of the $i$-th query node in each meta-task\\
$\mathbf{p}_i$&the classification probabilities of the $i$-th query node over $\mathcal{C}_b$\\
\hline
\end{tabular}
\vspace{-0.4cm}
\end{table}

	\subsection{Reproducibility}
	
In this section, we present the details on the reproducibility of our experiments. More specifically, we first elaborate on the implementation setting of our experiments. Then we introduce the required packages with the corresponding versions, followed by the experimental settings of baselines used in our main experiments. Finally, we provide details of datasets used in this paper.
	\subsubsection{Implementation of TENT}
	\label{appendix:implementation}
	Our framework TENT is implemented based on PyTorch~\cite{paszke2017automatic}.	We train our model on a single 16GB Nvidia V100 GPU. For the specific implementation setting, we set the number of training epochs $T$ as 500. We implement $\text{GNN}_\theta$ and $\text{GNN}_\phi$ using two-layer GINs~\cite{xu2018powerful} with the hidden sizes $d_h$ and $d_s$ both set as 16. To effectively initialize GNNs in our experiments, we utilize the Xavier initialization~\cite{glorot2010understanding}. The $\text{READOUT}$ function is implemented as mean-pooling. For the model
optimization, we adopt Adam~\cite{kingma2014adam} with the learning rate of 0.05
and a dropout rate of 0.2. The weight decay rate is set as $10^{-4}$ and the loss weight $\gamma$ is set as $1$. Finally, the model that achieves the best result on the validation dataset will be saved and used for test. In addition, we randomly sample 500 tasks from novel classes $\mathcal{C}_n$ (i.e., $T_{test}$=500) for test with a query set size $|\mathcal{Q}|$ of 10. Furthermore, to keep consistency, the test tasks are identical for all baselines. Our code is provided at \href{https://github.com/SongW-SW/TENT}{https://github.com/SongW-SW/TENT}.

	\subsubsection{Required Packages}
The more detailed package requirements are listed as below.
	\begin{itemize}
	    \item Python == 3.7.10
	    \item torch == 1.8.1
	    \item torch-cluster == 1.5.9
	    \item torch-scatter == 2.0.6
	    \item torch-sparse == 0.6.9
	    \item torch-geometric == 1.4.1
	    \item torch-spline-conv==1.2.1
	    \item numpy == 1.18.5
        \item scipy == 1.5.3
        \item cuda == 11.0
    \item tensorboard == 2.2.2
    \item networkx == 2.5.1
    \item scikit-learn == 0.24.1
    \item pandas==1.2.3
	\end{itemize}

	\subsubsection{Baseline Setting}
	Here, we present the detailed parameter setting of baselines. We mainly follow the original setting in the corresponding source code while adopting specific selections of parameters for better performance.
	\begin{itemize}
	    \item\textbf{Prototypical Network (PN)}~\cite{snell2017prototypical}: For PN, we set the learning rate as 0.005 with a weight decay of 0.0005. 
	            \item \textbf{MAML}~\cite{finn2017model}: The meta-learning rate is set as 0.001 and the number of update step is 10 with a learning rate of 0.01.
    \item \textbf{GCN}~\cite{kipf2017semi}: The learning rate is set as 0.001 and the hidden size of GCN is set as 32.
    \item \textbf{G-Meta}~\cite{huang2020graph}: For G-Meta, we set the meta-learning rate as 0.001. The number of update step is 10 and the update learning rate is 0.01. The dimension size of GNN is 128.
	    \item \textbf{GPN}~\cite{ding2020graph}: For GPN, we follow the setting in the source code and set the learning rate as 0.005 with a weight decay of 0.0005. The dimension sizes of two GNNs used in GPN are set as 32 and 16, respectively.
        \item \textbf{RALE}~\cite{liu2021relative}: We follow the setting in the source code and set the learning rates for training and fine-tuning as
0.001 and 0.01, respectively. The dropout rate is set as 0.6. The hidden size of used GNNs is 32.

	\end{itemize}
	
	\subsubsection{Dataset Description}
	In this section, we describe the detailed dataset settings. Specifically, among the four prevalent datasets used in our experiments, \texttt{Amazon-E}~\cite{mcauley2015inferring} and \texttt{DBLP}~\cite{tang2008arnetminer} datasets are obtained from~\cite{ding2020graph}, while \texttt{Cora-full}~\cite{bojchevski2018deep} and \texttt{OGBN-arxiv}~\cite{hu2020open} are obtained from the corresponding sources and processed by us. The statistics and details are as follows:
	\label{appendix}
	\begin{itemize}
    \item \textbf{Amazon-E}~\cite{mcauley2015inferring} is a product network, where nodes represent different "Electronics" products on Amazon. Moreover, edges are created according to the "viewed" relationship and class labels are assigned from the low-level product categories. For this dataset, we use 90/37/40 node classes for training/validation/test.
    \item \textbf{DBLP}~\cite{tang2008arnetminer} is a citation network. More specifically, each node represents a paper, and links are created according to the citation relations. The attributes are obtained via the paper abstract, and the class labels denote the paper venues. For this dataset, we use 80/27/30 node classes for training/validation/test.

    \item \textbf{Cora-full}~\cite{bojchevski2018deep} is a prevalent citation network, where nodes are labeled based on the paper topic. This dataset extends the prevalent small dataset via extracting original data from the entire network. For this
dataset, we use 25/20/25 node classes for training/validation/test.
    \item \textbf{OGBN-arxiv}~\cite{hu2020open} is a directed citation network of all CS arXiv papers indexed by MAG~\cite{wang2020microsoft}, where nodes represent arXiv papers and edges indicate citations. The feature of each node is a 128-dimensional feature vector obtained by averaging the embeddings of words in its title and abstract. The labels are assigned according to 40 subject areas of arXiv CS papers. We use 15/5/20 node classes for training/validation/test.
\end{itemize}

	\end{appendices}
	
\end{document}